\documentclass[pdflatex,sn-vancouver-num]{sn-jnl}

\usepackage{graphicx}%
\usepackage{multirow}%
\usepackage{amsmath,amssymb,amsfonts}%
\usepackage{amsthm}%
\usepackage{mathrsfs}%
\usepackage[title]{appendix}%
\usepackage{xcolor}%
\usepackage{textcomp}%
\usepackage{manyfoot}%
\usepackage{booktabs}%
\usepackage{algorithm}%
\usepackage{algorithmicx}%
\usepackage{algpseudocode}%
\usepackage{listings}%

\theoremstyle{thmstyleone}%
\theoremstyle{thmstyletwo}%

\theoremstyle{thmstylethree}%

\begin{document}

\title{A Diagnostic Evaluation of Neural Networks Trained with the Error Diffusion Learning Algorithm}

\author*[1]{\fnm{Kazuhisa} \sur{Fujita} \email{kazu@spikingneuron.net}}

\affil*[1]{\orgname{Komatsu University}, \orgaddress{\street{10-10 Doihara-Machi}, \city{Komatsu}, \postcode{923-0921}, \state{Ishikawa}, \country{Japan}}}

\abstract{
  The Error Diffusion Learning Algorithm (EDLA) is a learning scheme that performs synaptically local weight updates driven by a single, globally defined error signal. Although originally proposed as an alternative to backpropagation, its behavior has not been systematically characterized. We provide a modern formulation and implementation of EDLA and evaluate multilayer perceptrons trained with EDLA on parity, regression, and image-classification benchmarks (Digits, MNIST, Fashion-MNIST, and CIFAR-10). Following the original formulation, multi-class classification is implemented by training independent single-output networks (one per class), which makes the computational cost scale linearly with the number of classes. Under comparable architectures and training protocols, EDLA consistently underperforms backpropagation-trained baselines on all benchmarks considered. Through an analysis of internal dynamics, we identify a depth-related failure mode in ReLU-based EDLA: activations can grow explosively, causing unstable training and degraded accuracy. To mitigate this instability, we incorporate root mean square normalization (RMSNorm) into EDLA training. RMSNorm substantially improves numerical stability and expands the depth range in which EDLA can be trained, but it does not close the accuracy gap and retains the overhead of the parallel-network implementation. Overall, we offer a diagnostic evaluation of where and why global error diffusion breaks down in deep networks, providing guidance for future development of local, biologically inspired learning rules.
}


\maketitle

\section{Introduction}

Artificial neural networks have substantially advanced numerous research domains, including image processing, signal processing, and natural language processing. The remarkable success of these networks largely stems from the backpropagation method proposed by Rumelhart et al. \citep{Rumelhart:1986}, which has become a standard method for training neural networks across diverse architectures.

The theoretical foundation of neural networks traces back to the linear threshold unit (neuron) introduced by McCulloch and Pitts \citep{Kleene:1956,Mcculloch:1943}. The fundamental mechanism of this neuron is based on the all-or-none law \citep{Adrian:1914,Kato:1926} and still underlies modern neural network architectures. Rosenblatt's development of the perceptron \citep{Rosenblatt:1958} was another critical milestone in neural networks, marking the first realization of a trainable artificial neuron. However, because this perceptron consists of only a single computational layer, it cannot inherently form nonlinear decision boundaries. Although multilayer perceptrons (MLPs) resolve this limitation, training these deeper networks has long been a substantial challenge. The introduction of backpropagation provided a practical solution for training MLPs, and it subsequently became the dominant paradigm for training deep neural networks.

Despite the success of backpropagation, questions remain about its biological plausibility \citep{Crick:1989,Mostafa:2018}. Standard formulations require non-local credit assignment and tightly coordinated propagation of error information, which is difficult to reconcile with known neurobiological constraints. This gap between artificial and biological neural systems has driven the development of alternative learning algorithms, including approaches based on random feedback pathways (e.g., feedback alignment) \citep{Lillicrap:2016}, energy-based alternatives such as equilibrium propagation \citep{Scellier:2016}, and learning schemes that rely on local learning rules and/or local error signals \citep{Mostafa:2018}. 

In this context, Kaneko's Error Diffusion Learning Algorithm (EDLA) \citep{Kaneko} is a distinctive proposal that performs synaptically local weight updates driven by a single global error signal diffused throughout the network. EDLA introduces a biologically inspired architecture that incorporates both excitatory and inhibitory synapses. A distinctive feature of this algorithm is its error-diffusion mechanism: the discrepancy between the network output and target values is diffused throughout the network during training, and the original study reported that this can enable effective training with sigmoid activations. Unlike standard backpropagation, EDLA diffuses only a single error signal throughout the network, eliminating the need for explicit layer-wise backward error propagation. This simplification addresses a commonly noted limitation of backpropagation in biologically motivated settings \citep{Mostafa:2018}: standard training typically relies on storing forward activations and coordinating them with error information conveyed via a distinct backward pathway.

The original EDLA formulation lacks a mathematical representation consistent with modern deep learning standards. Furthermore, the behavior of EDLA in contemporary deep-network settings remains to be fully understood. In particular, the interaction between global error diffusion and modern components, such as deep architectures and unbounded activation functions, has not been systematically explored. The aim of this study is therefore to provide a formal modern description and a rigorous characterization of EDLA's capabilities and limitations in these settings. Rather than positioning EDLA as an immediate competitor to backpropagation, we use it as a case study to elucidate the dynamical constraints of global error signals in deep networks. We provide a modern implementation of EDLA and evaluate MLPs trained with EDLA on benchmark tasks including parity checks, regression, and image classification. We further analyze internal dynamics to understand the mechanisms of instability in deep rectified linear unit (ReLU) networks and assess whether standard stabilization interventions can extend the depth range in which EDLA can be trained reliably.

\paragraph{Contributions.}
The main contributions of this work are as follows:
\begin{itemize}
  \item We provide a modern formulation and implementation of the Error Diffusion Learning Algorithm (EDLA) and document the experimental setup and training protocol used throughout this study.
  \item We perform a controlled benchmark study of multilayer perceptrons trained with EDLA on parity, regression, and image-classification tasks (MNIST, Fashion-MNIST, CIFAR-10) under matched architectures and training settings, and we quantify the performance gap relative to standard backpropagation baselines.
  \item We make explicit the computational characteristics of the original EDLA multi-class formulation (independent single-output networks, one per class), including its linear scaling with the number of classes and its practical overhead compared to conventional multi-output MLPs.
  \item Through an analysis of internal network dynamics, we identify exploding activations as a primary failure mode of EDLA in deep ReLU networks and characterize the depth regime in which training becomes unstable.
  \item We evaluate Root Mean Square normalization (RMSNorm) as a stabilization intervention, showing that it improves numerical stability and extends the trainable depth range, while not eliminating the accuracy gap or the parallel-network overhead.
\end{itemize}

\section{Error Diffusion Learning Algorithm (EDLA)}

EDLA is a learning scheme proposed by Kaneko \citep{Kaneko}. Although the original model is available only in Japanese, this section provides a comprehensive reinterpretation of EDLA.

\subsection{Network Architecture}

\begin{figure}
 \begin{center}
 \includegraphics[width=0.5\linewidth]{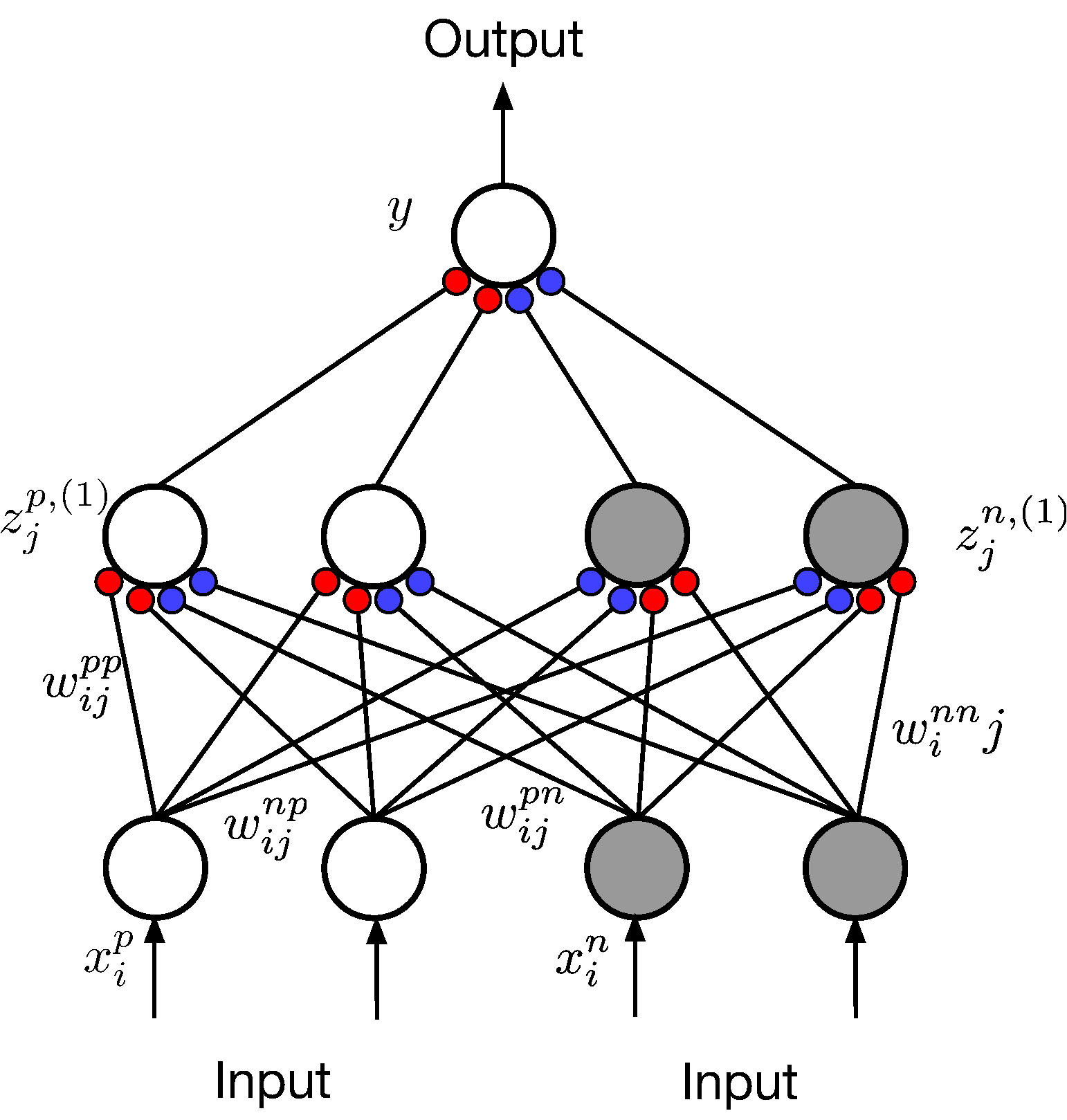}
 \caption{Schematic representation of the Error Diffusion Learning Algorithm (EDLA) network architecture. The network consists of multiple layers of positive neurons (white circles) and negative neurons (gray circles). The red and blue small circles at the connections indicate excitatory and inhibitory synapses, respectively.}
 \label{fig:edla}
 \end{center}
\end{figure}

Figure \ref{fig:edla} illustrates the neural network architecture employed by the EDLA. Like traditional MLPs, EDLA adopts a multilayer structure comprising neurons and synapses. However, EDLA incorporates two types of neurons: positive neurons (p-neurons) and negative neurons (n-neurons). The synaptic connections are categorized based on the types of neurons they link: excitatory synapses connect neurons of identical type, whereas inhibitory synapses connect neurons of different types. A key characteristic of the EDLA is that it has only one output neuron.

\subsection{Learning Rule and Mathematical Formulation}

The input to the EDLA network is represented by the vector $\mathbf{x} = [x_0, x_1, \ldots, x_N]^T \in \mathbb{R}^{N+1}$, where $x_0$ corresponds to the bias set to one, and $N$ is the number of input features. In the EDLA architecture, each layer consists of positive and negative sublayers. The outputs from the positive and negative sublayers of the input layer are denoted by $\mathbf{x}^p$ and $\mathbf{x}^n$, respectively. For the input layer, $\mathbf{x}^p = \mathbf{x}^n = \mathbf{x}$.

For a neuron $j$ in the first hidden layer $(l=1)$, the activations for the positive and negative neurons, $a_j^{p,(1)}$ and $a_j^{n,(1)}$, are computed as follows:
\begin{equation}
a_j^{p,(1)} = {\mathbf{w}_j^{pp,(1)}}^T \mathbf{x}^p + {\mathbf{w}_j^{pn,(1)}}^T \mathbf{x}^n,
\end{equation}
\begin{equation}
a_j^{n,(1)} = {\mathbf{w}_j^{np,(1)}}^T \mathbf{x}^p + {\mathbf{w}_j^{nn,(1)}}^T \mathbf{x}^n.
\end{equation}
In these equations, $\mathbf{w}_j^{pp,(1)} = [w_{j0}^{pp}, \ldots, w_{jN}^{pp}]^T$ represents the vector of synaptic weights connecting the positive neurons in the input layer to positive neuron $j$ in the first hidden layer $(l=1)$. Similarly, $\mathbf{w}_j^{pn,(1)}$, $\mathbf{w}_j^{np,(1)}$, and $\mathbf{w}_j^{nn,(1)}$ denote the weight vectors for the other combinations of neuron types in the connections between the input and the first hidden layers. Here, $pn$ and $np$ indicate connections from negative to positive neurons and from positive to negative neurons, respectively.

The outputs of the positive and negative neurons $j$ in the first hidden layer are given by:
\begin{equation}
z_j^{p, (1)} = g(a_j^{p, (1)}) \quad \text{and} \quad z_j^{n, (1)} = g(a_j^{n, (1)}),
\end{equation}
where $g(\cdot)$ is the activation function. In subsequent hidden layers $(l = 2, ..., L)$ where $L$ refers to the final hidden layer, the neuron outputs are given by
\begin{equation}
z_j^{p, (l)} = g({\mathbf{w}_j^{pp, (l)}}^T \mathbf{z}^{p,(l-1)} + {\mathbf{w}_j^{pn, (l)}}^T \mathbf{z}^{n,(l-1)}),
\end{equation}
\begin{equation}
z_j^{n, (l)} = g({\mathbf{w}_j^{np, (l)}}^T \mathbf{z}^{p,(l-1)} + {\mathbf{w}_j^{nn, (l)}}^T \mathbf{z}^{n,(l-1)}).
\end{equation}
The final output of the EDLA network $y$ is computed as
\begin{equation}
y = g({\mathbf{w}^{pp, (L+1)}}^T \mathbf{z}^{p,(L)} + {\mathbf{w}^{pn, (L+1)}}^T \mathbf{z}^{n, (L)}).
\end{equation}

Synaptic weights within the EDLA network are constrained to maintain specific relationships, reflecting the excitatory and inhibitory nature of the connections:
\begin{equation}
w^{pp, (l)}_{ji} \geq 0, \quad w^{pn, (l)}_{ji} \leq 0, \quad w^{np, (l)}_{ji} \leq 0, \quad w^{nn, (l)}_{ji} \geq 0.
\end{equation}
To reflect these constraints at the start of training, weights are initialized by sampling uniformly from $[0,1]$ for synapses connecting neurons of the same type and from $[-1,0]$ for those connecting neurons of different types. For the non-negative activation functions considered in this study (sigmoid and ReLU), these sign constraints are preserved by the EDLA update rule because both the presynaptic activities and the derivative terms are non-negative (since $z \geq 0$ and $g'(a) \geq 0$). Therefore, we do not apply explicit clipping or projection in our implementation.

The objective of training an EDLA network is to identify the optimal parameter set $\mathbf{W}$ that minimizes a predefined loss function $E(\mathbf{W})$, similar to traditional MLPs. Gradient descent is a widely used optimization method in which parameters are iteratively updated in a direction that locally reduces the value of $E(\mathbf{W})$. In practice, most practitioners use stochastic gradient descent (SGD) \citep{Lecun:2015}, which computes gradient estimates efficiently using subsets of training data (mini-batches). The EDLA also employs a method based on SGD.

Although the EDLA utilizes the principles of gradient descent, it incorporates an approximation in computing the \emph{error signals} (often called \emph{error terms} in the classical literature \citep{Bishop:2007,Bishop:2024}) used for weight updates. In an MLP trained by backpropagation, the error signal $\delta_j^{(l)}$ for neuron $j$ in hidden layer $l$ is computed as:
\begin{equation}
\delta_j^{(l)} = g'(a_j^{(l)}) \sum_{k} w_{kj}^{(l+1)} \delta_k^{(l+1)},
\end{equation}
where $g'(a_j^{(l)})$ is the derivative of the activation function. Here, $\delta_j^{(l)}$ denotes a backpropagated error signal (local sensitivity), i.e., $\delta_j^{(l)} \equiv \partial E(\mathbf{W})/\partial a_j^{(l)}$. With this definition, the parameter gradient is given by $\partial E(\mathbf{W})/\partial w_{kj}^{(l+1)} = \delta_k^{(l+1)} z_j^{(l)}$, which yields the usual gradient-descent update.

In contrast, rather than computing exact local sensitivities via the recursive sum $\sum_{k} w_{kj}^{(l+1)} \delta_k^{(l+1)}$, the EDLA approximates the update direction using a single global error signal $d$, defined as
\begin{equation}
d = - \frac{\partial E(\mathbf{W})}{\partial y}.
\end{equation}
This global error signal, whose sign dictates which synapses to update, is uniformly diffused throughout the network. Specifically, the approximate local update signals, $\tilde{\delta}^{pp, (l)}_{j}$, $\tilde{\delta}^{np, (l)}_{j}$, $\tilde{\delta}^{nn, (l)}_{j}$, and $\tilde{\delta}^{pn, (l)}_{j}$ in the EDLA, are expressed as:
\begin{align}
\nonumber \tilde{\delta}^{pp, (l)}_{j} = d g'(a^{p, (l)}_j), & \tilde{\delta}^{np, (l)}_{j} = d g'(a^{n, (l)}_j),\\
\tilde{\delta}^{nn, (l)}_{j} = d g'(a^{n, (l)}_j), & \tilde{\delta}^{pn, (l)}_{j} = d g'(a^{p, (l)}_j).
\end{align}

A key distinction between EDLA and traditional backpropagation is EDLA's simplified approximation of error computations. Whereas traditional backpropagation individually computes and propagates neuron-specific error signals $\delta$s from subsequent layers, EDLA diffuses only one global error signal $d$ uniformly across all layers. This diffusion mechanism significantly reduces complexity by eliminating the need for individual neuron-level error calculations, highlighting a unique feature of the EDLA approach compared with traditional backpropagation methods.

The weight updates in EDLA differ depending on the sign of the global error signal $d$. When $d > 0$, the weights originating from the positive neurons $w^{pp, (l)}_{ji}$ and $w^{np, (l)}_{ji}$ are updated as follows:
\begin{align}
w^{pp, (l)}_{ji} \leftarrow w^{pp, (l)}_{ji} + \eta \tilde{\delta}^{pp, (l)}_{j} z^{p, (l-1)}_i
= w^{pp, (l)}_{ji} + \eta d g'(a^{p, (l)}_j) z^{p, (l-1)}_i,\\
w^{np, (l)}_{ji} \leftarrow w^{np, (l)}_{ji} - \eta \tilde{\delta}^{np, (l)}_{j} z^{p, (l-1)}_i
= w^{np, (l)}_{ji} - \eta d g'(a^{n, (l)}_j) z^{p, (l-1)}_i.
\end{align}
Conversely, when $d < 0$, the weights originating from the negative neurons, $w^{nn, (l)}_{ji}$ and $w^{pn, (l)}_{ji}$ are updated as:
\begin{align}
w^{nn, (l)}_{ji} \leftarrow w^{nn, (l)}_{ji} - \eta \tilde{\delta}^{nn, (l)}_{j} z^{n, (l-1)}_i
= w^{nn, (l)}_{ji} - \eta d g'(a^{n, (l)}_j) z^{n, (l-1)}_i,\\
w^{pn, (l)}_{ji} \leftarrow w^{pn, (l)}_{ji} + \eta \tilde{\delta}^{pn, (l)}_{j} z^{n, (l-1)}_i
= w^{pn, (l)}_{ji} + \eta d g'(a^{p, (l)}_j) z^{n, (l-1)}_i.
\end{align}
where $\eta$ denotes the learning rate. Thus, the global error signal directly determines whether weights from positive or negative neurons are adjusted during training.

The behavior of the EDLA becomes intuitively clear when employing the sigmoid activation function $g(a) = 1/(1 + \exp(-a))$ along with the squared error loss function defined as $E(\mathbf{W}) = \frac{1}{2}(y - t)^2$. The global error signal $d$ is represented as follows:
\begin{equation}
d = - \frac{\partial E(\mathbf{W})}{\partial y} = - \frac{\partial}{\partial y}\left[\frac{1}{2}(y - t)^2\right] = t - y.
\end{equation}
When the network output is smaller than the target value ($d > 0$), the absolute values of the weights originating from positive neurons increase in magnitude when $g'(a)> 0$ and $z > 0$. Such weight adjustments amplify the outputs of positive neurons in subsequent layers while simultaneously suppressing the outputs of negative neurons in those layers. Consequently, these modifications reduce the discrepancy between the output and target value of the network. Conversely, when the network output exceeds the target value ($d < 0$), the absolute values of the weights originating from negative neurons increase. These weight adjustments decrease the outputs of positive neurons in subsequent layers while simultaneously increasing the outputs of negative neurons. Thus, regardless of the sign of the global error signal, EDLA systematically reduces the difference between the network output and target values through successive weight updates. Importantly, this characteristic learning behavior of EDLA is maintained as long as $g'(a) \geq 0$, $z \geq 0$, and the sign of the derivative of the loss function properly encodes whether the network output is below or above the target value. The same squared error loss function was used in all experiments of EDLA conducted in this study.

For the EDLA, the activation function is typically chosen as the sigmoid function. It constrains the output of the neuron between 0 and 1 and its derivative between 0 and 0.25. By limiting both the neuron outputs and their derivatives, the sigmoid activation function provides stability to the training dynamics, avoiding large weight updates and facilitating stable convergence.

It is worth noting that if the derivative term $g'(a)$ were replaced by the postsynaptic activity $z=g(a)$, the update rule would reduce to a standard Hebbian form. In practice, this modification can lead to unstable dynamics because the update magnitude scales directly with activity, which may allow runaway weight growth when activations become large. The original EDLA formulation instead employs $g'(a)$ as an activity-dependent gain. For the sigmoid activation, $g'$ approaches zero in saturated regimes, thereby diminishing updates for highly active units and helping to prevent runaway weight dynamics. In contrast, for ReLU, $g'(a) \in \{0, 1\}$ lacks this saturation mechanism, which motivates the stabilization analyses for deep ReLU networks presented in later sections.

\subsection{Biological Interpretation}

In the EDLA framework, the global error signal $d$ functions as a plasticity-guiding neuromodulatory signal, orchestrating synaptic weight updates through the coordinated activities of presynaptic and postsynaptic neurons. This concept is strongly reminiscent of Hebbian learning \citep{Hebb:1949}, which posits that changes in synaptic strength are driven by correlated patterns of neuronal activation. Specifically, the EDLA learning rule, written as $\eta d g'(a^{p, (l)}_j) z^{p, (l-1)}_i$ for synaptic weights $w^{pp,(l)}_{ji}$, is proportional to the product of the postsynaptic activation derivative $g'(a^{p, (l)}_j)$ and the presynaptic output $z^{p, (l-1)}_i$. This mathematical relationship can therefore be interpreted as a Hebbian-like plasticity mechanism: weight updates reflect the interaction between a presynaptic neuron (characterized by $z^{p, (l-1)}_i$) and a postsynaptic neuron (captured by $g'(a^{p, (l)}_j)$).

Moreover, EDLA employs a global error-diffusion mechanism in which the error signal $d$ is broadcast across the network. This diffusion is conceptually analogous to biological neuromodulation and closely parallels reward-prediction-error (RPE) signals observed in biological systems, which guide synaptic plasticity via diffuse neuromodulatory pathways (e.g., dopamine) \citep{Roelfsema:2018}, thereby mirroring how neuromodulators broadly regulate synaptic change.

Unlike traditional Hebbian learning—which is typically unsupervised and driven solely by local activity correlations—EDLA integrates a supervised component through the global error signal $d$, modulating synaptic changes according to network performance relative to a target. Conceptually, EDLA aligns with neo-Hebbian three-factor rules in which synaptic change depends on (i) a presynaptic factor, (ii) a postsynaptic factor, and (iii) a systems-level modulatory factor \citep{Lisman:2011,Fremaux:2016}. In EDLA the error $d$ plays this third role: weight updates depend on presynaptic activity $z^{p, (l-1)}_i$, a postsynaptic state $g'(a^{p, (l)}_j)$, and the global error signal $d$. 

In canonical neuroscientific three-factor formulations, pre- and postsynaptic coincidence first sets a synapse-specific eligibility trace (or tag), and a later modulatory signal (e.g., dopamine/RPE) converts that trace into an actual weight update. Eligibility traces bridge the temporal gap between neural activity and delayed reward signals, which is particularly important in spiking networks where modulatory signals often arrive with delays relative to the activity that made a synapse ``eligible'' \citep{Fremaux:2016}. By contrast, EDLA operates in an artificial-network regime where the global error signal is computed and broadcast immediately at each training step. Accordingly, EDLA may be viewed as a limiting case of three-factor rules in which eligibility traces have effectively negligible lifetime and modulatory delays are absent; this approximation is reasonable for supervised tasks with immediate error availability but limits applicability in scenarios with delayed modulatory feedback.

\subsection{Structural Constraints of Global Error Diffusion}

The inherent simplicity and approximation underlying EDLA introduce significant constraints, particularly concerning the network's capability to handle multiple outputs effectively. In the EDLA framework, the global error signal $d$ is uniformly diffused throughout the network architecture. This global error diffusion inherently directs the optimization process toward minimizing the loss associated with a single-output node. Consequently, when dealing with single-output scenarios, the EDLA can efficiently and accurately specialize the network weights for a specific output. However, when an EDLA network contains multiple outputs, each output generates its own global error signal, which propagates throughout the network. Such simultaneous diffusion of distinct error signals leads to conflicting weight adjustments, as the network simultaneously attempts to minimize losses for multiple outputs. These conflicting updates inhibit the network's ability to extract and generalize meaningful features from the input data, thereby undermining learning efficacy.

\subsection{Extending EDLA to Multiple Outputs}
\label{sec:multi_output}

The inherent single-output constraint of the EDLA represents a significant limitation when addressing tasks that require multiple outputs. A simple approach to adapting EDLA to overcome this limitation involves constructing multiple EDLA networks arranged in parallel. This parallelized architecture addresses a task that requires $K$ outputs by employing $K$ independent EDLA networks. Each individual network retains the fundamental EDLA architecture and operates autonomously. Input vector $\mathbf{x}$ is simultaneously provided to all networks, ensuring that they process the same input data. Each network independently computes output $y_k$, where $k \in \{1, \ldots, K\}$. In classification tasks, if the output $y_k$ is the maximum value among the outputs of the EDLA networks, $k$ is regarded as the overall output label.

The weights of each individual EDLA network are independently optimized using the standard EDLA update rule. For multi-class classification, each single-output EDLA network $k$ is trained independently with the squared-error loss
$E_k=\frac{1}{2}(y_k-t_k)^2$, where $t_k\in\{0,1\}$ is a one-vs-rest target. Accordingly, the global error signal for network $k$ is $d_k=-\partial E_k/\partial y_k = t_k-y_k$. Consequently, this parallel architecture retains the capability, simplicity, and core principles inherent to single-output EDLA models, effectively extending EDLA's capability to handle multi-output tasks. However, this parallelization approach increases computational costs because the total number of EDLA networks scales linearly with the required number of outputs. This trade-off highlights the challenge of maintaining the computational simplicity of single-output EDLA networks while contending with the additional resource demands necessary for multi-output learning.

Employing multiple single-output networks in an EDLA framework may seem computationally inefficient compared with traditional multi-output MLP architectures. Nevertheless, this parallelization method is reminiscent of modular structures, which are organizational principles observed in biological neural systems, particularly in the brain \citep{Casanova:2018}. Mountcastle's early neurophysiological research \citep{Mountcastle:1957} has demonstrated that neurons within the primate somatosensory cortex are organized into vertically aligned columns, each dedicated to a particular modality or receptive field. Additionally, neurons in the primary visual cortex are organized into modular structures known as hypercolumns, each specializing in processing specific features of visual stimuli such as orientation \citep{Hubel:1968}. Each hypercolumn functions semi-independently, collectively contributing to overall visual perception by handling distinct components of the visual scene. Similarly, the inferior temporal cortex exhibits clusters of neurons that are selectively responsive to complex visual features including distinct object shapes \citep{Sato:2008,Tanaka:1996}. This clustered arrangement enables the efficient processing of diverse visual elements while simultaneously facilitating effective integration across features and categories. Another notable example is the barrel cortex of rodents, in which neural clusters (called barrels) are dedicated to independently processing tactile inputs from individual whiskers \citep{Li:2011,Woolsey:1970}. Each barrel independently processes sensory information corresponding to its associated whisker, and the combined neural activity across the barrels generates a comprehensive tactile representation. These modular and distributed neural processing strategies enable the brain to handle complex and heterogeneous tasks efficiently while preserving both flexibility and robustness. Although the parallelized architecture of EDLA, which consists of multiple single-output networks dedicated to specific outputs, may be computationally less efficient than traditional multi-output MLPs, it reflects these biological modular architectures.

The increased computational cost of the parallelized EDLA is an important practical consideration. However, the increase may be less problematic for brain systems or neuromorphic computing applications. In such systems, neurons and synapses are implemented as components that operate individually and execute in massive parallelism \citep{Sharp:2014,Akopyan:2015}. As a result, increases in neuron and synapse counts do not necessarily translate into proportional increases in wall-clock computation time. Some areas of the brain discussed above, such as the visual cortex and the barrel cortex, employ modular architectures that process complex sensory inputs in real-time. Indeed, researchers have achieved real-time simulation using massively parallel systems \citep{Sharp:2014}. Therefore, while the parallelized EDLA entails a trade-off between biological plausibility and computational efficiency on standard computer systems, it may be particularly well suited to neuromorphic implementations, brain-inspired computing systems, or other massively parallel platforms. However, the increase in the number of unit circuits for neuromorphic systems and the memory cost for parallel computing remain a concern, as each EDLA network requires its own set of weights and parameters.

\subsection{Parameter Complexity and Scalability Analysis}
\label{sec:param_count}

For an EDLA network with $L$ hidden layers, $n_{\mathrm{hid}}$ positive neurons and $n_{\mathrm{hid}}$ negative neurons per layer, input size $n_{\mathrm{in}}$, and $n_{\mathrm{out}}$ outputs (implemented as $n_{\mathrm{out}}$ independent single-output EDLA networks), the total number of parameters, $N_{\mathrm{EDLA}}$, is
\begin{align}
N_{\mathrm{EDLA}}
&= \big(2 n_{\mathrm{in}} (2 n_{\mathrm{hid}}) + (L-1)(2n_{\mathrm{hid}})^2 + 2 n_{\mathrm{hid}}\big)\, n_{\mathrm{out}}\\
&= 2 n_{\mathrm{out}}\, n_{\mathrm{hid}} \big(2 n_{\mathrm{in}} + 2 (L-1) n_{\mathrm{hid}} + 1\big).
\end{align}
Here the first term, $2 n_{\mathrm{in}} (2 n_{\mathrm{hid}})$, corresponds to the weights connecting the input layer to the first hidden layer; the second term, $(L-1)(2n_{\mathrm{hid}})^2$, accounts for the weights between hidden layers; and the third term, $2 n_{\mathrm{hid}}$, corresponds to the weights from the last hidden layer to a single output (there are $n_{\mathrm{hid}}$ positive and $n_{\mathrm{hid}}$ negative neurons, hence $2 n_{\mathrm{hid}}$). Multiplying by $n_{\mathrm{out}}$ yields the total number of parameters across all independent single-output EDLA networks. Bias terms are omitted in this theoretical analysis for simplicity. The exact parameter counts reported in Table \ref{tab:performance_selected}, which are used for experimental comparison, include bias terms.

In contrast, the parameter count for a standard fully connected MLP with $L$ hidden layers, $m_{\mathrm{hid}}$ neurons per hidden layer, input size $n_{\mathrm{in}}$, and $n_{\mathrm{out}}$ outputs, $N_{\mathrm{MLP}}$, is
\begin{align}
N_{\mathrm{MLP}} &= n_{\mathrm{in}} m_{\mathrm{hid}} + (L-1) m_{\mathrm{hid}}^2 + m_{\mathrm{hid}} n_{\mathrm{out}}\\
&= m_{\mathrm{hid}}\big(n_{\mathrm{in}} + (L-1) m_{\mathrm{hid}} + n_{\mathrm{out}}\big),
\end{align}
again omitting bias terms. Unlike EDLA, the parameter count of a traditional MLP does not scale strongly with $n_{\mathrm{out}}$ (it appears only in the final layer term).

To compare parameter efficiency, consider the case $n_{\mathrm{hid}} = m_{\mathrm{hid}}$ and assume the network is sufficiently deep and wide so that hidden-to-hidden weights dominate. When $2 (L-1) n_{\mathrm{hid}} \gg 2 n_{\mathrm{in}} + 1$, the EDLA parameter count can be approximated as
\begin{align}
N_{\mathrm{EDLA}} \approx 4\, n_{\mathrm{out}}\, n_{\mathrm{hid}}^2 (L-1).
\end{align}
If $n_{\mathrm{hid}} = m_{\mathrm{hid}}$ and $(L-1) m_{\mathrm{hid}} \gg n_{\mathrm{in}} + n_{\mathrm{out}}$, then
\begin{align}
N_{\mathrm{MLP}} \approx n_{\mathrm{hid}}^2 (L-1).
\end{align}
Thus, in this limit,
\begin{align}
\frac{N_{\mathrm{EDLA}}}{N_{\mathrm{MLP}}} \approx 4\, n_{\mathrm{out}}.
\end{align}
An EDLA implemented as $n_{\mathrm{out}}$ independent single-output networks can therefore require roughly $4\,n_{\mathrm{out}}$ times more parameters than a comparable MLP with the same hidden-layer size and depth. This substantial increase in the parameter count can raise memory usage and computational cost during training and inference, particularly for tasks with many outputs. The computational costs may, however, be partially mitigated by parallel implementations or neuromorphic hardware, as noted in the previous subsection. If a multi-output EDLA architecture were developed that shares computations across outputs in future work, it could substantially reduce the parameter overhead compared with the current parallelized approach.

\subsection{Stabilization Strategy: RMSNorm}
\label{sec:rms}

When the activation function has no upper bound (for example, ReLU), individual pre-activation values and the corresponding outputs can grow without bound. Consequently, the magnitude of weight updates depends directly on the activation of the preceding layer. In particular, large pre-activations under ReLU can produce very large updates and thereby destabilize the learning dynamics. To mitigate this problem, Root Mean Square (RMS) normalization (RMSNorm) \citep{Zhang:2019} is applied to the pre-activations. Normalization techniques such as RMS, Layer Normalization \citep{Ba:2016}, and batch normalization \citep{Ioffe:2015} are commonly applied to pre-activations in artificial neural networks.

The RMS is defined as
\begin{equation}
\mathrm{RMS} = \sqrt{\frac{1}{2 n_{\mathrm{hid}}}\sum_{j=1}^{n_{\mathrm{hid}}}\left( (a^{p, (l)}_j)^2 + (a^{n, (l)}_j)^2 \right)}
\end{equation}
where $n_{\mathrm{hid}}$ is the number of neurons in each sublayer (so that $2 n_{\mathrm{hid}}$ is the total number of hidden neurons in the full layer). The normalized pre-activation $\tilde{a}^{p, (l)}_{j}$ is then given by
\begin{equation}
\tilde{a}^{p, (l)}_{j} = \frac{a^{p, (l)}_j}{\mathrm{RMS}} \gamma.
\end{equation}
An analogous expression holds for $\tilde{a}^{n,(l)}_{j}$. The positive scalar $\gamma$ controls the overall magnitude of the normalized pre-activations. In this study, $\gamma$ is fixed to 1.0 because learning $\gamma$ is outside the scope of this work and because $\gamma$ is here treated as a biologically motivated gain parameter. A fixed $\gamma$ scales the pre-activation vector of each layer without changing its direction. RMSNorm is applied to the pre-activations of all layers except the output layer, whereas the input remains unnormalized prior to the first linear transform.

Incorporating RMSNorm, the weight update rule for a connection from a positive presynaptic neuron to a positive postsynaptic neuron is derived as follows:
\begin{align}
  \label{eq:dw_rmsnorm}
dw^{pp, (l)}_{ji} &= \eta d g'(\tilde{a}^{p, (l)}_{j}) \frac{\partial \tilde{a}^{p, (l)}_{j}}{\partial a^{p, (l)}_j} z^{p, (l-1)}_i\\
&= \eta d g'(\tilde{a}^{p, (l)}_{j}) \frac{1}{\mathrm{RMS}}\left(1 - \frac{{a^{p, (l)}_j}^2}{2 n_{\mathrm{hid}} \mathrm{RMS}^2} \right) z^{p, (l-1)}_i
\end{align}
The term $1 - \frac{{a^{p, (l)}_j}^2}{2 n_{\mathrm{hid}} \mathrm{RMS}^2}$ is non-negative because $2 n_{\mathrm{hid}} \mathrm{RMS}^2$ is the sum of all squared pre-activations in the layer, which is necessarily greater than any single squared pre-activation $(a^{p, (l)}_j)^2$. Therefore, crucially, the sign of the weight update $dw^{pp, (l)}_{ji}$ remains determined by the sign of the global error signal $d$, preserving the fundamental learning mechanism of the EDLA. The update rules for other weight types are derived in an analogous manner.

RMSNorm can be interpreted in the context of biological information processing, relating to mechanisms like gain control and divisive normalization. Gain control \citep{Fujita:2007,Mehaffey:2005} modulates neuronal sensitivity according to the overall activity level of a population, thereby helping to maintain firing rates within a stable dynamic range and avoiding saturation. Divisive normalization \citep{Carandini:2011,Heeger:1992} divides a neuron's response by a pooled measure of population activity, preventing any single unit from dominating the representation. RMSNorm performs a related operation by scaling pre-activations in proportion to the population RMS; this reduces the magnitude of excessively large updates and thus stabilizes learning.

RMSNorm is selected over Layer Normalization due to a key property of the EDLA architecture. In EDLA, non-negative activation functions such as the sigmoid or ReLU are frequently employed. Layer Normalization standardizes activations to have zero mean and unit variance, a process that involves re-centering the data. This re-centering inevitably shifts many pre-activation values into the negative range. In particular, when using the ReLU activation function, this would cause a significant portion of neurons to output zero, effectively silencing them and impeding the weight updates. RMSNorm avoids this issue by rescaling activations without the re-centering step.

Note that RMSNorm is applied only in the experiments reported in Section~\ref{sec:stabilization}; it is omitted in the other experiments described in this manuscript.

Integrating the core learning rule, error diffusion mechanism, and the optional RMSNorm stabilization strategy described above, we summarize the overall training procedure of the EDLA for a single-output network in Algorithm \ref{alg:edla_single}. We use mini-batches; Algorithm 1 describes the per-sample update, and in practice we compute updates per sample and aggregate them by mean over the batch (batch size reported in Table 2). For compactness we use a concatenated (positive/negative) notation in hidden layers; at the output layer we follow the original single-output EDLA and compute/update only the positive unit.

\begin{algorithm}[t]
\caption{EDLA: a single output and one-sample update (optional RMSNorm)}
\label{alg:edla_single}
\begin{algorithmic}[1]
\Require input $x\in\mathbb{R}^{n_{\mathrm{in}}}$, target $t\in\mathbb{R}$ \Comment{$x_0=1$ absorbs bias as in the paper}
\Require parameters $\{W^{(l)}_p, W^{(l)}_n\}_{l=1}^{L+1}$, where $L$ is the number of hidden layers
\Statex \Comment{$W^{(l)}_p=[W^{pp,(l)};W^{np,(l)}],\;W^{(l)}_n=[W^{pn,(l)};W^{nn,(l)}]$ (row-wise concat)}
\Require activations $\{g_l\}_{l=1}^{L+1}$ and derivatives $\{g_l'\}_{l=1}^{L+1}$, learning rate $\eta$
\Require (optional) RMSNorm enabled/disabled (paper: used only for $1\le l\le L$)

\Statex
\Function{NormAndDiagJac}{$a,l$}
  \If{(RMSNorm disabled) \textbf{or} $l=L+1$}
     \State $\tilde a \gets a$; \;\; $\rho \gets \mathbf{1}$
  \Else \Comment{$1\le l\le L$}
     \State $\tilde a \gets \mathrm{RMSNorm}(a)$
     \State $\rho \gets \text{element-wise factor } \rho_j = \frac{\partial \tilde a_j}{\partial a_j}\; \text{(diagonal of the Jacobian)}$
  \EndIf
  \State \Return $(\tilde a,\rho)$
\EndFunction

\Statex
\Statex \textbf{Forward (single sample)}
\State $z^{(0)}_{p} \gets x$;\;\; $z^{(0)}_{n} \gets x$
\For{$l=1$ to $L+1$}
  \State $a^{(l)} \gets W^{(l)}_p z^{(l-1)}_{p} \;+\; W^{(l)}_n z^{(l-1)}_{n}$
  \Statex \hspace{1em}\Comment{$a^{(l)}=[a^{p,(l)},a^{n,(l)}]\in\mathbb{R}^{2n_l}$ (concat of pos/neg neurons)}
  \State $(\tilde a^{(l)}, \rho^{(l)}) \gets$ \Call{NormAndDiagJac}{$a^{(l)},l$}
  \State $z^{(l)} \gets g_l(\tilde a^{(l)})$ \Comment{$z^{(l)}=[z^{p,(l)},z^{n,(l)}]$}
  \State $\phi^{(l)} \gets g_l'(\tilde a^{(l)}) \odot \rho^{(l)}$
\EndFor
\State $y \gets z^{p,(L+1)}$ \Comment{output is the positive half of the final layer}
\State $d \gets t - y$ \Comment{global error signal}
\State $d^{+} \gets \max(d,0)$;\;\; $d^{-} \gets \max(-d,0)$

\Statex
\Statex \textbf{Weight update (error diffusion; single sample)}
\For{$l=1$ to $L+1$}
  \State $\Delta W^{(l)}_p \gets \eta \, d^{+}\, \big(\phi^{(l)} {z^{(l-1)}_{p}}^{\!\top}\big)\odot \mathrm{sign}(W^{(l)}_p)$
  \State $\Delta W^{(l)}_n \gets \eta \, d^{-}\, \big(\phi^{(l)} {z^{(l-1)}_{n}}^{\!\top}\big)\odot \mathrm{sign}(W^{(l)}_n)$
  \State $W^{(l)}_p \gets W^{(l)}_p + \Delta W^{(l)}_p$
  \State $W^{(l)}_n \gets W^{(l)}_n + \Delta W^{(l)}_n$
\EndFor
\end{algorithmic}
\end{algorithm}

\section{Experimental Setup}

The neural networks are implemented in Python using NumPy for linear algebra operations and PyTorch for neural networks. The source code is available on GitHub \url{https://github.com/KazuhisaFujita/EDLA} for transparency and further extension of this work.

\section{Experimental Results}

This section evaluates the performance of the proposed EDLA network across various tasks. To provide a clear and comprehensive overview of the experimental environment, Table \ref{tab:dataset_summary} summarizes the characteristics of the benchmark datasets, including input dimensions, preprocessing steps, and train/test data splits. Furthermore, Table \ref{tab:experimental_setup} outlines the key hyperparameter settings, loss functions, and network architectures used for both EDLA and the baseline MLPs. Detailed descriptions and analyses of the results are provided in the subsequent subsections (Section \ref{sec:regression} for regression and Section \ref{sec:image_classification} for image classification).

\begin{table*}[htbp]
\caption{Summary of benchmark datasets, preprocessing steps, and train/test splits used in the experiments.}
\label{tab:dataset_summary}
\centering
\resizebox{\textwidth}{!}{%
\begin{tabular}{@{}llcccl@{}}
\toprule
\textbf{Dataset} & \textbf{Task Type} & \textbf{Total} & \textbf{Input Dim.} & \textbf{Preprocessing} & \textbf{Train/Test} \\
\midrule
Airfoil Self-Noise & Regression & 1,503 & 5 & MinMax Scaling & 80\% / 20\% (Random) \\
Concrete Strength  & Regression & 1,030 & 8 & MinMax Scaling & 80\% / 20\% (Random) \\
Energy Efficiency  & Regression & 768   & 8 & Target: Heating Load only, MinMax Scaling & 80\% / 20\% (Random) \\
\midrule
Digits             & Classification & 1,797   & $8 \times 8$ & MinMax Scaling, Flatten & 80\% / 20\% (Random) \\
MNIST              & Classification & 70,000 & $28 \times 28$ & Normalized $[0, 1]$, Flatten & 60,000 / 10,000 (Fixed) \\
Fashion-MNIST      & Classification & 70,000 & $28 \times 28$ & Normalized $[0, 1]$, Flatten & 60,000 / 10,000 (Fixed) \\
CIFAR-10           & Classification & 60,000 & $32 \times 32 \times 3$ & Normalized $[0, 1]$, Flatten & 50,000 / 10,000 (Fixed) \\
\bottomrule
\end{tabular}%
}
\end{table*}

\begin{table*}[htbp]
\caption{Summary of experimental setups, architectures, and hyperparameters. Evaluated network depths are $L \in \{1, 2, 4\}$ hidden layers, with varying widths. RMSNorm is evaluated for EDLA-ReLU as an optional stabilizer (EDLA (ReLU+RMS)). The initial scale (IS) for weights is 1.0 by default, and 0.0001 for small-scale initialization EDLA with ReLU AF. The MLP baseline is trained using AdamW (weight decay = 0) with a constant learning rate.}
\label{tab:experimental_setup}
\centering
\resizebox{\textwidth}{!}{%
\begin{tabular}{@{}lll c c l@{}}
\toprule
\textbf{Model} & \textbf{Loss Function} & \textbf{Activation (Hidden $\to$ Out)} & \textbf{Learning Rate} & \textbf{Batch} & \textbf{Epochs} \\
\midrule
\multicolumn{6}{c}{\textbf{Regression Tasks}} \\
\midrule
EDLA &  MSE & Sigmoid $\to$ Identity & 0.001  & 64 & 500 \\
EDLA &  MSE & ReLU    $\to$ Identity & 0.0001 & 64 & 500 \\
MLP  &  MSE & Sigmoid $\to$ Identity & 0.1    & 64 & 500 \\
MLP  &  MSE & ReLU    $\to$ Identity & 0.01   & 64 & 500 \\
\midrule
\multicolumn{6}{c}{\textbf{Image Classification Tasks}} \\
\midrule
EDLA &  MSE           & Sigmoid $\to$ Sigmoid  & 1.0    & 128 & 500 (Digits/MNIST/Fashion), 1000 (CIFAR-10) \\
EDLA &  MSE           & ReLU    $\to$ Sigmoid  & 0.1    & 128 & 500 (Digits/MNIST/Fashion), 1000 (CIFAR-10) \\
MLP  &  Cross-Entropy & Sigmoid $\to$ Identity & 0.001  & 128 & 500 (Digits/MNIST/Fashion), 1000 (CIFAR-10) \\
MLP  &  Cross-Entropy & ReLU    $\to$ Identity & 0.0001 & 128 & 500 (Digits/MNIST/Fashion), 1000 (CIFAR-10) \\
\bottomrule
\end{tabular}%
}
\end{table*}

\subsection{Parity Check Task}

We first evaluate EDLA on a parity-check task as a sanity check in a controlled low-dimensional setting. The task is binary classification: given a binary input vector, the model predicts whether the number of ones is odd or even. We report $n_{\mathrm{bit}}=5$ as a representative and challenging setting, and use the complete set of $2^{n_{\mathrm{bit}}}$ input patterns. EDLA uses a single sigmoid output neuron whose value is interpreted as $p(\text{odd})$, with a decision threshold of 0.5. A mini-batch size of 4 is used throughout (fixed across all parity settings) for consistency with the $n_{\mathrm{bit}}=2$ case (where the full dataset contains only $2^{2}=4$ patterns), and results are averaged over repeated runs (see figure captions). Results for $n_{\mathrm{bit}}\in\{2,3,4\}$ show the same qualitative trends (high accuracy with sigmoid activations and learning-rate sensitivity with ReLU) and are provided in the Supplementary Material (Figs.~S1--S4).

Figure~\ref{fig:parity_sigmoid_summary} summarizes performance with sigmoid hidden activations. EDLA reaches near-perfect accuracy for most widths, with small-width settings showing a modest drop depending on LR/depth. Increasing width generally reduces the epochs required to reach 0.9 accuracy, indicating faster convergence with higher capacity. For moderate-to-large widths, learning rates around 1.0--10.0 converge substantially faster than 0.1. With a fixed learning rate (bottom row), increasing depth does not prevent convergence, although the time to reach the threshold can vary with depth and width.

Figure~\ref{fig:parity_relu_summary} summarizes performance with ReLU hidden activations. For moderate learning rates (e.g., 0.01--0.1), EDLA attains high accuracy and reaches the 0.9 threshold rapidly as width increases. In contrast, an overly large learning rate (LR=1.0) degrades performance and fails to reach the threshold for this setting, indicating sensitivity to step size when activations are unbounded. With a fixed learning rate (bottom row), deeper ReLU networks can remain accurate in this small task but may exhibit slower or less consistent convergence. This learning-rate sensitivity foreshadows the instability observed in deeper EDLA-ReLU networks on larger benchmarks, motivating the stabilization analysis in Sec.~\ref{sec:stabilization}.

\begin{figure}[htbp]
 \centering
 \includegraphics[width=0.95\linewidth]{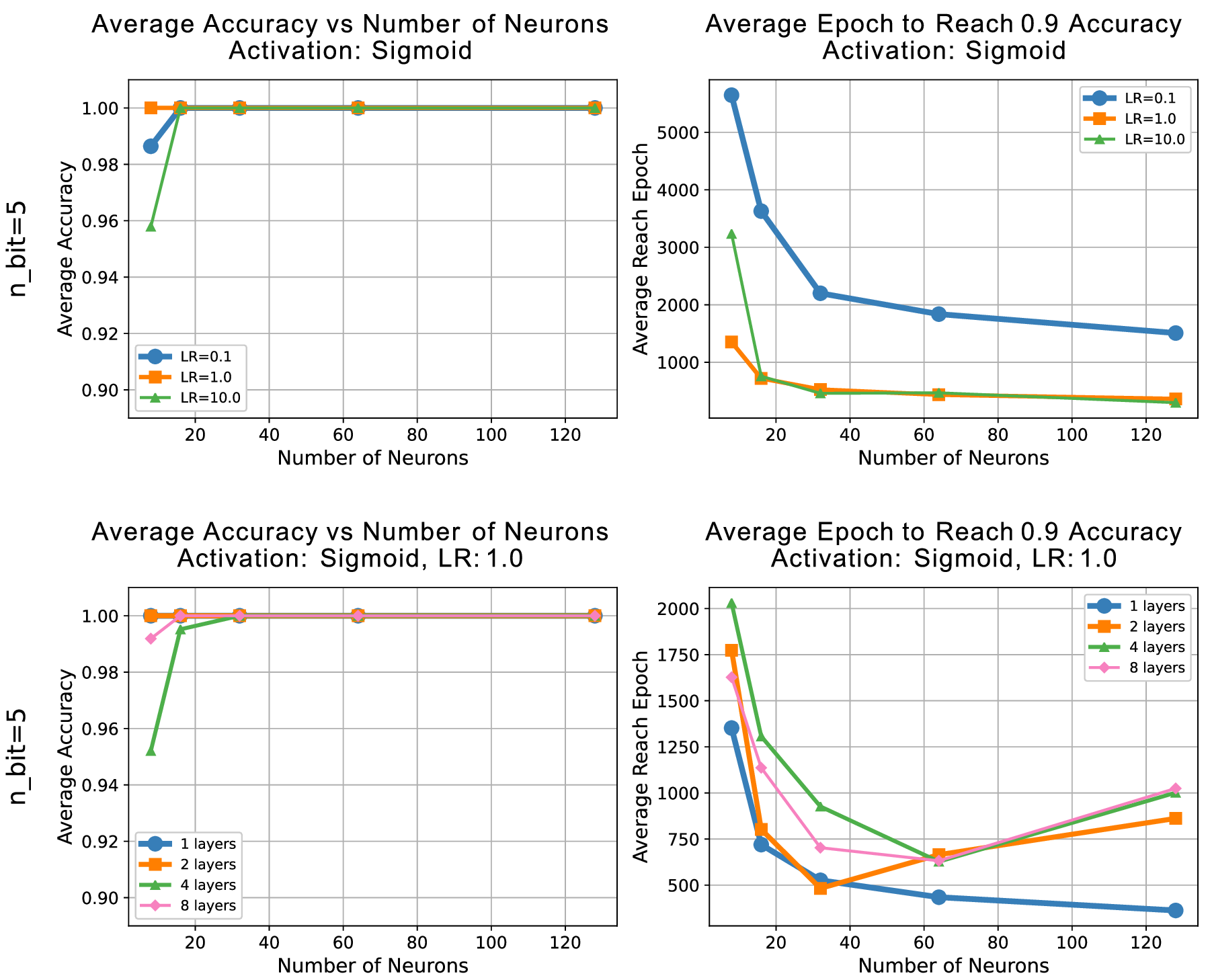}
 \caption{Parity-check results for EDLA with sigmoid hidden activations ($n_{\mathrm{bit}}=5$). 
Top row: learning-rate sweep for a 1-hidden-layer network (LR = 0.1, 1.0, 10.0). 
Bottom row: depth sweep (1, 2, 4, 8 hidden layers) at a fixed learning rate (LR = 1.0). 
Left panels report the final training accuracy, computed as the mean accuracy over epochs 19900--20000. Right panels report the mean epoch to 0.9 accuracy (computed over trials that reached the threshold) over 10 independent trials. All results are averaged over 10 independent trials using specific random seeds (48835, 52642, 7841, 58416, 96828, 34439, 25155, 52094, 23535, 49704).}
 \label{fig:parity_sigmoid_summary}
\end{figure}

\begin{figure}[htbp]
 \centering
 \includegraphics[width=0.95\linewidth]{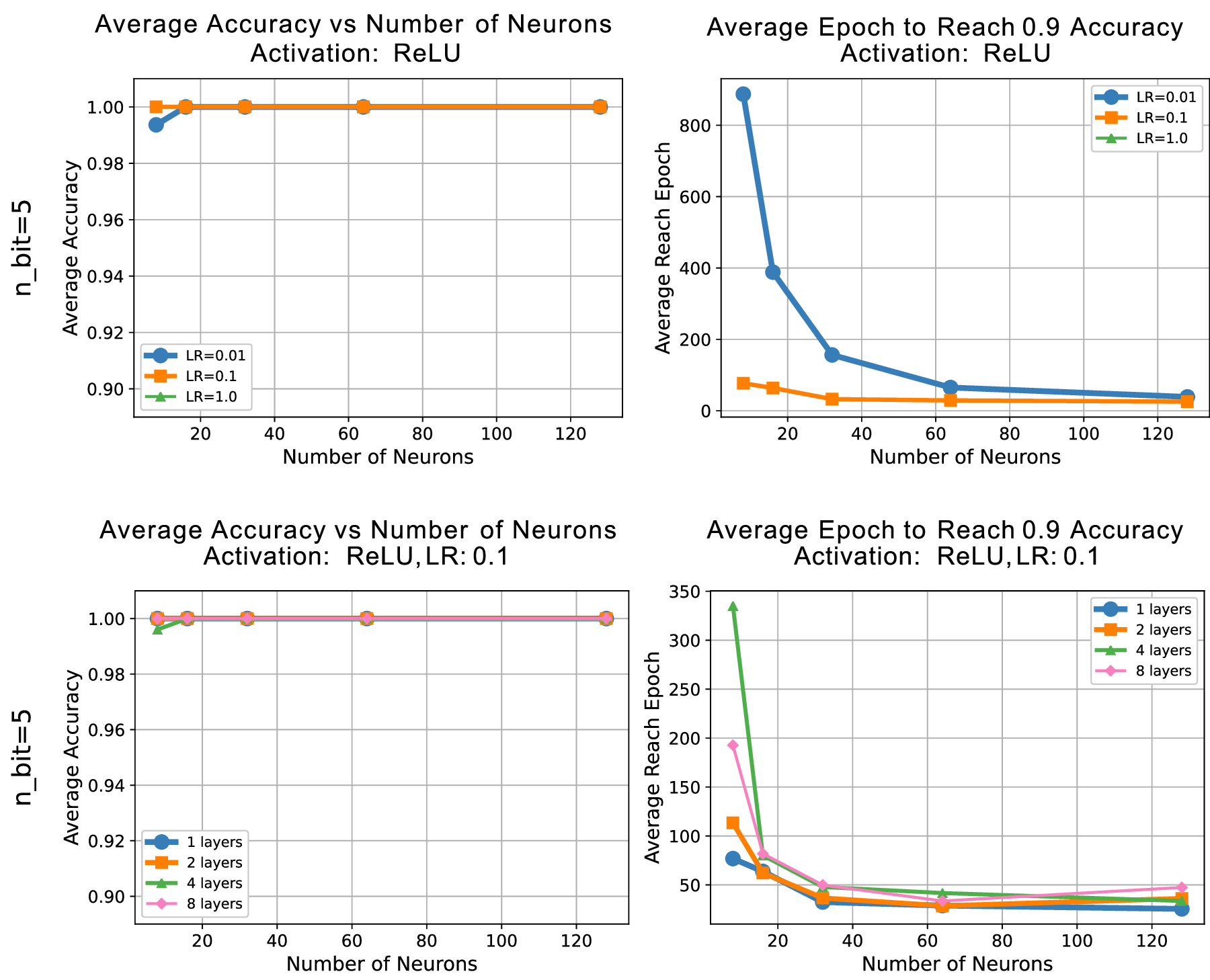}
\caption{Parity-check results for EDLA with ReLU hidden activations ($n_{\mathrm{bit}}=5$). 
Top row: learning-rate sweep for a 1-hidden-layer network (LR = 0.01, 0.1, 1.0). 
Bottom row: depth sweep (1, 2, 4, 8 hidden layers) at a fixed learning rate (LR = 0.1). 
Left panels report the final training accuracy, computed as the mean accuracy over epochs 19900--20000. In the left panels, curves may be absent when the final accuracy falls below the plotted y-axis range. Right panels report the mean epoch to 0.9 accuracy (computed over trials that reached the threshold). Missing points in the right panels indicate that no trials reached 0.9 within 20000 epochs. All results are averaged over 10 independent trials, employing the same random seeds used in Fig.~\ref{fig:parity_sigmoid_summary} to ensure consistency.}
 \label{fig:parity_relu_summary}
\end{figure}

\subsection{Regression Tasks}
\label{sec:regression}

The EDLA network is evaluated on regression tasks using three benchmark datasets: Airfoil Self-Noise, Concrete Compressive Strength, and Energy Efficiency. The Airfoil Self-Noise dataset comprises 1503 samples with five features and a target variable representing the sound pressure level. The Concrete Compressive Strength dataset includes 1030 samples with eight features and a target variable denoting the compressive strength of concrete. The Energy Efficiency dataset comprises 768 samples with eight features and two target variables: heating and cooling loads. In this experiment, only the heating load is employed as the target variable, and the cooling load is not considered. The training and test datasets are randomly split into 80\% and 20\%, respectively, for all datasets.

The EDLA network consists of an input layer, hidden layers, and an output layer. The neurons in the hidden layers use the sigmoid or ReLU activation functions (AFs). The neurons in the output layer use the identity AF (i.e., no AF). Initially, we applied the learning rates identified in the parity-check task (1.0 and 0.1 for sigmoid and ReLU, respectively); however, these relatively large rates resulted in lower performance in the regression setting. Consequently, we performed a small learning-rate sweep using the Concrete Compressive Strength dataset to determine task-appropriate values. We evaluated learning rates of $\eta \in \{0.0001, 0.001, 0.01\}$ for the sigmoid AF and $\eta \in \{0.00001, 0.0001, 0.001\}$ for the ReLU AF. Based on the lowest Mean Absolute Error (MAE), the learning rates were set to 0.001 for the sigmoid AF and 0.0001 for the ReLU AF. Training is conducted for 500 epochs employing a mini-batch size of 64 samples.

\begin{figure}
 \begin{center}
 \includegraphics[width=1\linewidth]{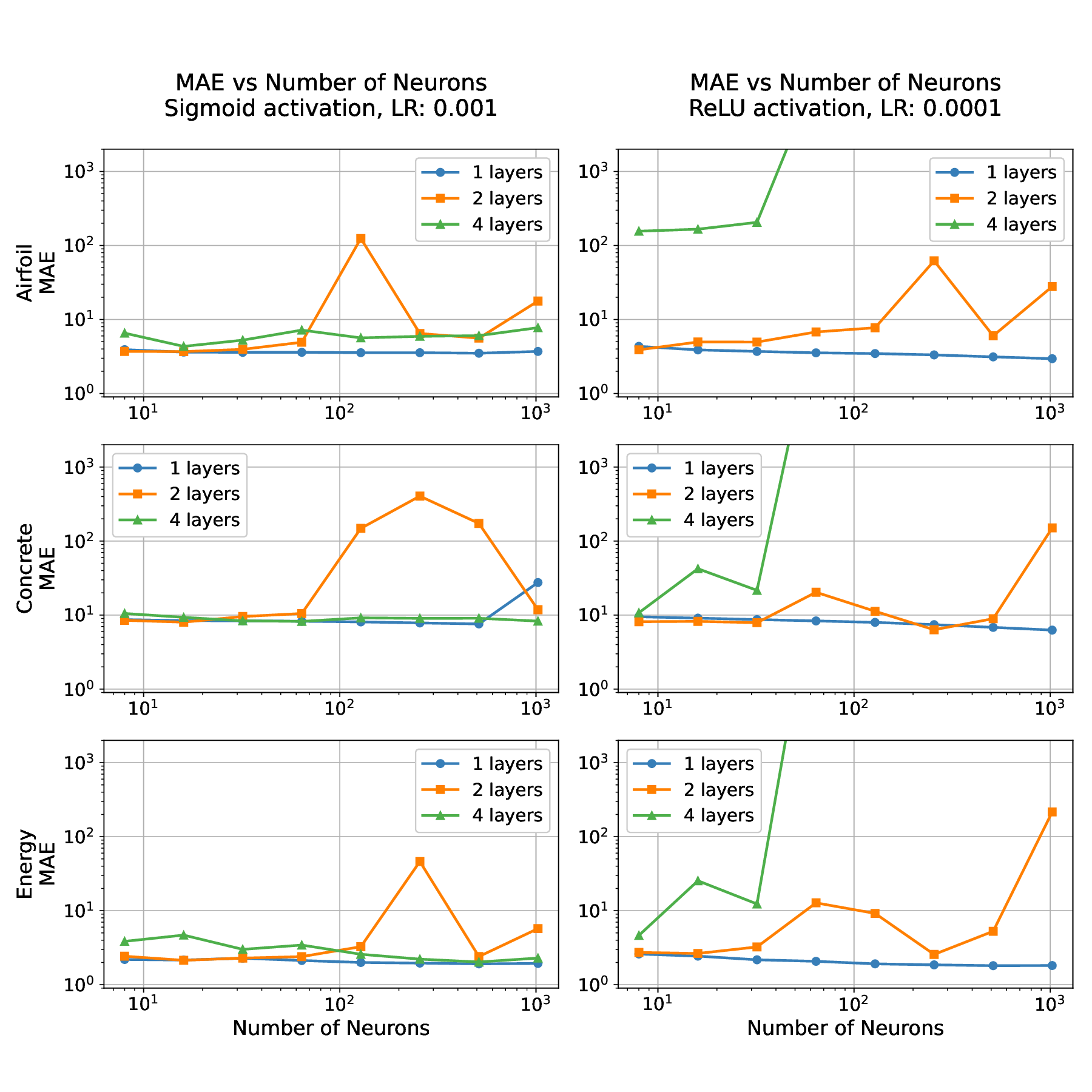}
 \caption{Performance of the EDLA network on regression tasks across varying layer configurations. Mean Absolute Error (MAE) is shown against the number of neurons per hidden layer for networks with 1, 2, and 4 hidden layers. The rows correspond to the Airfoil Self-Noise (top), Concrete Compressive Strength (middle), and Energy Efficiency (bottom) datasets. The left column shows results for networks using the sigmoid AF (learning rate = 0.001), while the right column shows results for ReLU activation (learning rate = 0.0001). Each point represents the average of four independent trials using specific random seeds (73313, 97895, 15503, 4387). Missing data points for the 4-layer ReLU network indicate training instability, where the MAE became excessively large or a NaN (not-a-number) value. The omitted data points (number of neurons, MAE) are as follows: Airfoil: (64, $1.87 \times 10^4$), (128, $4.74 \times 10^{14}$), (256, $1.93 \times 10^{31}$), (512, NaN), (1024, NaN); Concrete: (64, $1.06 \times 10^5$), (128, $8.45 \times 10^{15}$), (256, $1.45\times 10^{32}$), (512, NaN), (1024, NaN); Energy: (64, $4.44 \times 10^5$), (128, $2.03 \times 10^{16}$), (256, $1.63\times 10^{25}$), (512, NaN), (1024, NaN).}
 \label{fig:edla_regression}
 \end{center}
\end{figure}

Figure \ref{fig:edla_regression} illustrates the Mean Absolute Error (MAE) performance of the EDLA network across varying numbers of neurons (8, 16, 32, 64, 128, 256, 512, and 1024) and different layer configurations (1, 2, and 4 layers). The results indicate that EDLA networks with Sigmoid AF are generally stable and yield relatively low MAE for most widths, with occasional degradations at extreme widths (e.g., Concrete at 1024 neurons). However, the EDLA network with the sigmoid AF does not exhibit performance improvements as the number of neurons or network depth increases.

For the 1-layer networks, ReLU is comparable to or slightly better than sigmoid at moderate-to-large widths, and shows a mild improvement with width. In contrast, increasing the network depth does not yield performance improvements but rather leads to performance degradation compared with shallower architectures. This deterioration is likely caused by the instability of the EDLA network when using ReLU, due to the increase in both neuron count and network depth. As detailed in the caption (Figure \ref{fig:edla_regression}), this instability was particularly severe for the 4-layer ReLU network, resulting in severe divergence (off-scale MAE) or NaN values for several configurations.

These observations collectively suggest that, while the EDLA approach is adequately capable of handling regression tasks, augmenting the number of neurons and network depth does not necessarily improve performance. Furthermore, the EDLA network with the ReLU AF tends to exhibit inferior performance with increased depth. In the most stable (1-layer) configurations, ReLU can show a slight advantage over sigmoid.

\subsection{Image Classification Tasks}
\label{sec:image_classification}

To evaluate the performance of the EDLA network in image classification tasks, the experiments are conducted using the Digits dataset (Optical Recognition of Handwritten Digits), MNIST, Fashion-MNIST, and CIFAR-10 datasets. The Digits dataset consists of 1797 grayscale images with a resolution of $8 \times 8$ pixels. Because the Digits dataset is not divided into training and test datasets, 20 \% of the entire dataset is randomly selected and used as test data. The MNIST comprises 60,000 training and 10,000 testing grayscale images of handwritten digits, each with a resolution of $28 \times 28$ pixels. Similarly, the Fashion-MNIST includes 60,000 training and 10,000 testing grayscale images of Fashion articles at a resolution of $28 \times 28$ pixels. The CIFAR-10 consists of 50,000 training and 10,000 testing color images with a resolution of $32 \times 32$ pixels. The CIFAR-10 dataset is more complex than the other three datasets because it contains color images. In this experiment, the CIFAR-10 dataset is converted to a flattened one-dimensional vector format.

To handle these 10-class classification tasks, the overall architecture consists of 10 independent single-output EDLA networks trained in parallel, as detailed in Section \ref{sec:multi_output}. Each individual network consists of an input layer, hidden layers, and an output layer. The neurons in the hidden layers use either the sigmoid or ReLU activation functions. The output layer employs the sigmoid AF. To assess their performance across various configurations, the number of neurons in each hidden layer is varied systematically. The learning rates when using the sigmoid and ReLU AFs are 1.0 and 0.1, respectively. These learning rates were selected based on the 5-bit parity-check results (Figs.~\ref{fig:parity_sigmoid_summary} and \ref{fig:parity_relu_summary}), where they demonstrated the most reliable convergence. The networks are trained for 500 epochs when using the Digits, MNIST, and Fashion-MNIST datasets, and for 1000 epochs when using the CIFAR-10 dataset. A mini-batch size of 128 samples is used.

\begin{figure}
 \begin{center}
 \includegraphics[width=0.9\linewidth]{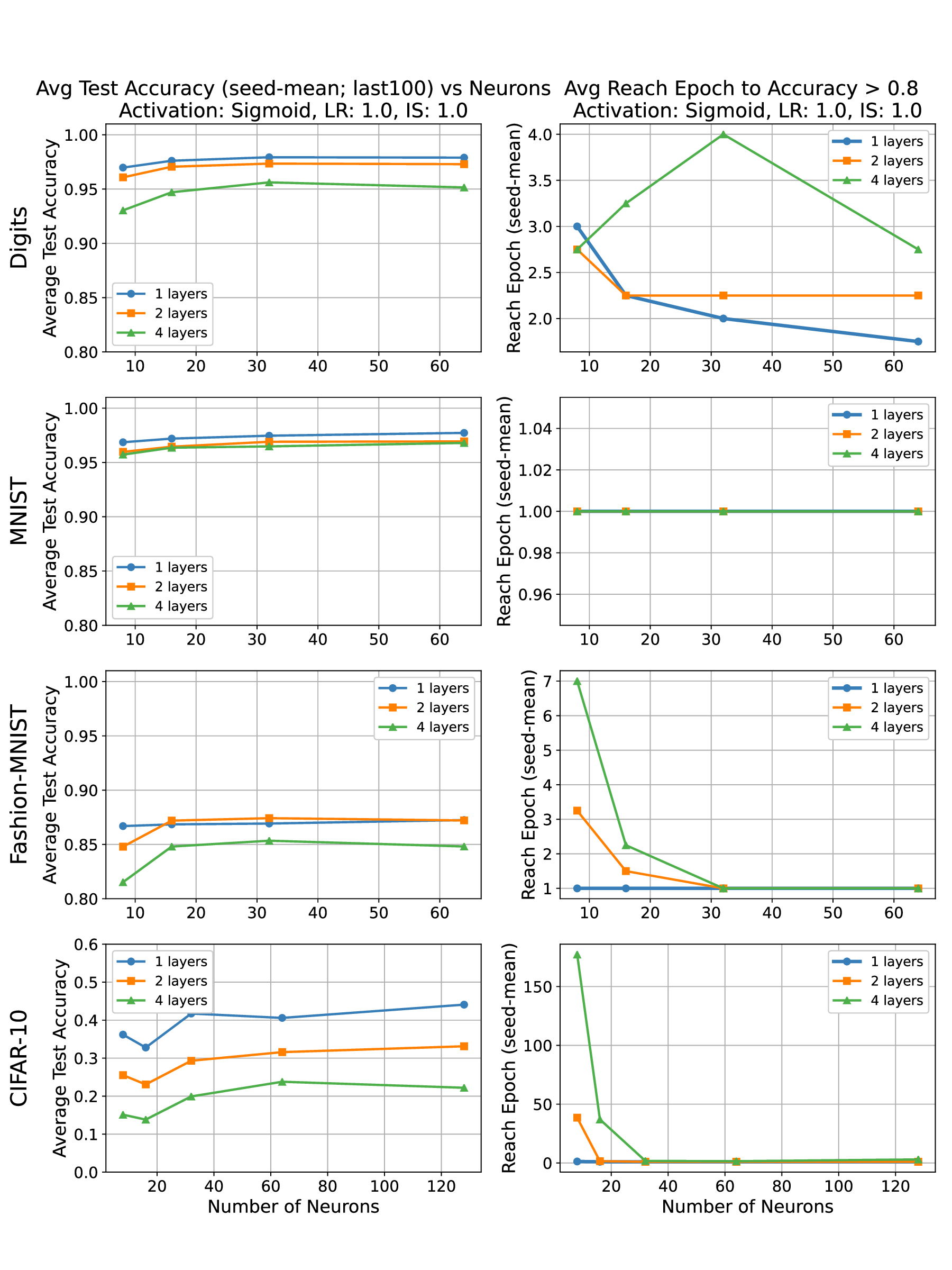}
 \caption{Performance of EDLA with the sigmoid AF on image datasets across layer configurations. The plots in the left column show the accuracy of the EDLA network with the sigmoid AF as a function of the number of neurons in each hidden layer for different layer configurations (1, 2, and 4 layers) across four datasets: Digits, MNIST, Fashion-MNIST, and CIFAR-10. The plots in the right column illustrate the number of epochs required to achieve an accuracy of $\theta_\mathrm{threshold}$ against the number of neurons in each hidden layer for the same layer configurations. For CIFAR-10,  $\theta_\mathrm{threshold}$ is set to 0.2. Otherwise, $\theta_\mathrm{threshold}$ is set to 0.8. The results are averaged over four independent trials using specific random seeds (40323, 52036, 34802, 31402).}
 \label{fig:edla_sigmoid}
 \end{center}
\end{figure}

Figure \ref{fig:edla_sigmoid} illustrates the performance of the EDLA network using the sigmoid AF for various numbers of neurons in each hidden layer. Across datasets, performance tends to improve with width. Specifically, for the Digits and MNIST datasets, near-perfect accuracy is rapidly achieved, even with a relatively small number of neurons. For the Fashion-MNIST dataset, while high accuracy is attained, a larger number of neurons contributed to incrementally improved performance. For the CIFAR-10 dataset, the performance is lower compared to the other datasets. This lower performance of CIFAR-10 is attributed to the network architecture, which has a simple layer structure and is not well-suited for processing complex color images as in convolutional neural networks. This width-dependent gain is most apparent for the shallow CIFAR-10 models, whereas deeper CIFAR-10 networks remain substantially less accurate. Moreover, across all datasets, the depth of the EDLA network does not improve performance and can even degrade it. Most configurations reach the threshold in a small number of epochs; notably, deeper and/or smaller-width CIFAR-10 models can require substantially more epochs. Increasing width typically speeds convergence, though the trend is not strictly monotonic.

\begin{figure}
 \begin{center}
 \includegraphics[width=0.9\linewidth]{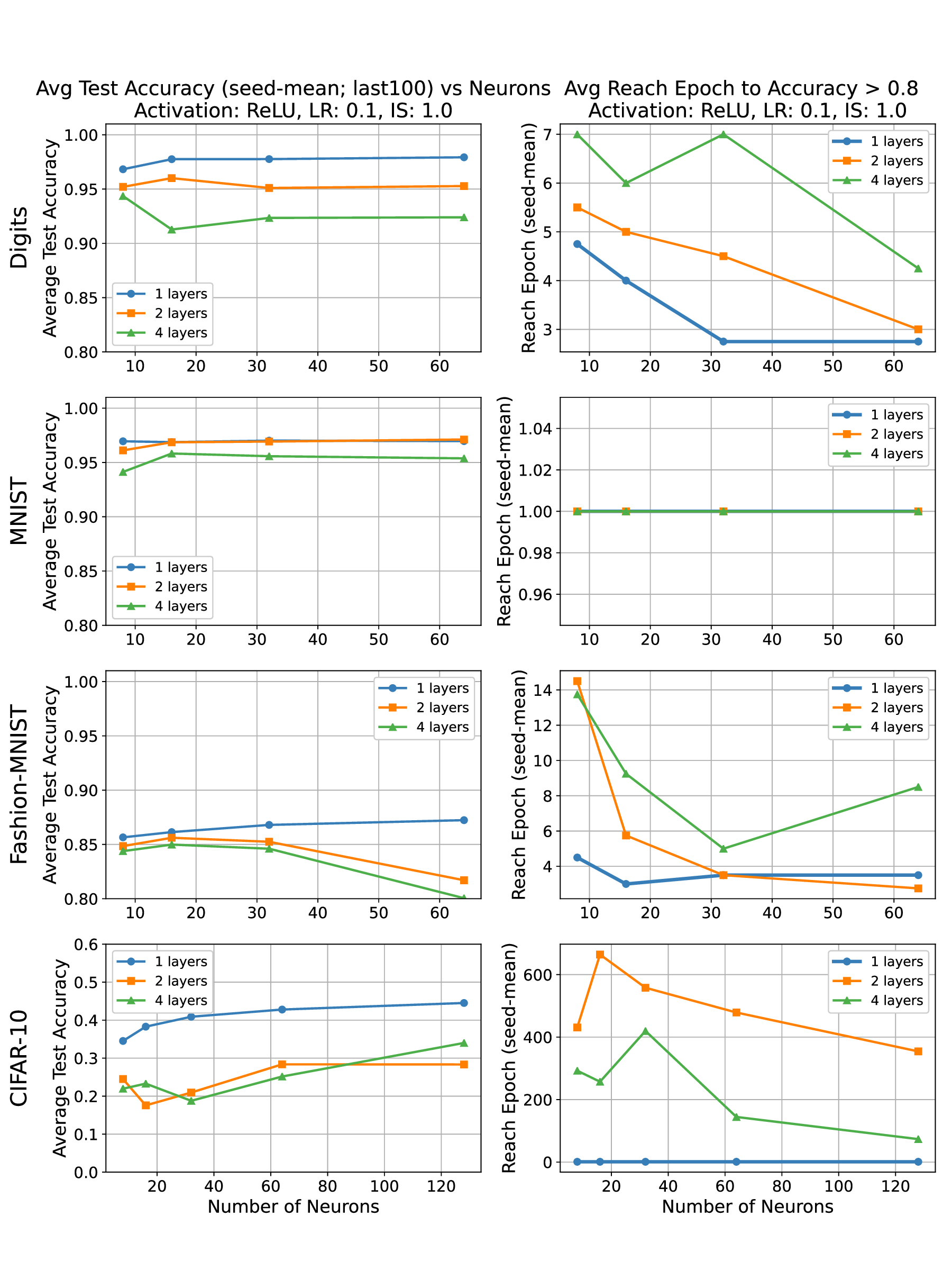}
 \caption{Performance of EDLA with the ReLU AF on image datasets across layer configurations. The plots in the left column show the accuracy of the EDLA network with the ReLU AF as a function of the number of neurons in each hidden layer for different layer configurations (1, 2, and 4 layers) across four datasets: Digits, MNIST, Fashion-MNIST, and CIFAR-10. The plots in the right column illustrate the number of epochs required to achieve an accuracy of $\theta_\mathrm{threshold}$ against the number of neurons in each hidden layer for the same layer configurations. For CIFAR-10,  $\theta_\mathrm{threshold}$ is set to 0.2. Otherwise, $\theta_\mathrm{threshold}$ is set to 0.8. The results are averaged over four independent trials, employing the same random seeds used in Fig.~\ref{fig:edla_sigmoid} to ensure consistency.}
 \label{fig:edla_relu}
 \end{center}
\end{figure}

Similarly, Figure \ref{fig:edla_relu} shows the performance of the EDLA network using the ReLU AF across varying numbers of neurons in each hidden layer. The results reveal that the EDLA network with the ReLU AF achieves a performance comparable to that of the sigmoid-based network. However, similar to the sigmoid network, increasing network depth generally leads to worse performance. The trend of accuracy versus neuron count is also more complex: while performance on CIFAR-10 generally improves with more neurons, performance on Digits (4-layer) and Fashion-MNIST (deeper networks) slightly degrades. The reach epoch often decreases with width for many settings (notably CIFAR-10), but can be non-monotonic for some datasets/depths. ReLU typically requires more epochs to reach the threshold than sigmoid (except on MNIST, where all settings reach the threshold almost immediately).

These observations collectively suggest that EDLA networks can effectively handle relatively simple image classification tasks using both sigmoid and ReLU activation functions. However, more complex datasets such as CIFAR-10 present challenges for the simple EDLA architecture employed in this study. Increasing the number of neurons in each hidden layer generally enhances performance, while increasing network depth does not necessarily yield improvements and may even degrade performance. Regarding convergence speed, increasing width typically accelerates convergence—most clearly for CIFAR-10, although the trend can be non-monotonic in some ReLU settings. Conversely, increasing network depth tended to slow convergence for sigmoid networks. Furthermore, the reach epoch for the ReLU AF was generally larger (slower) than that for the sigmoid AF across most configurations.

\subsection{Analysis of Internal Representations}

In this subsection, we investigate the internal representations learned by the EDLA network by analyzing the synaptic weights in the hidden layers. The experiments use an EDLA network with a single hidden layer containing two neurons with sigmoid AFs. The network is trained on the Digits dataset over 500 epochs.

\begin{figure}
 \centering
 \includegraphics[width=0.8\linewidth]{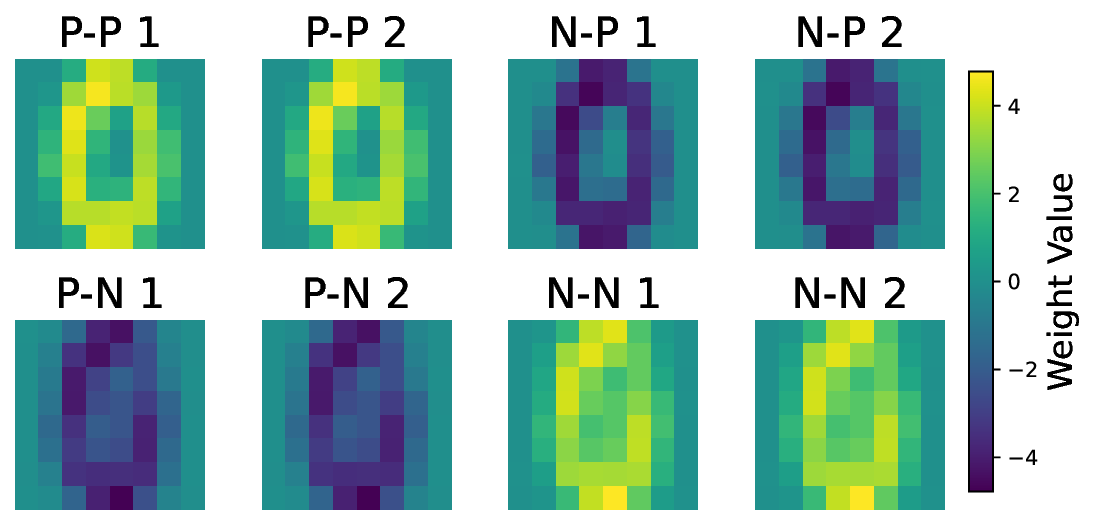}
 \caption{A heatmap of the synaptic weights connecting the input layer to the first hidden layer in the EDLA network, trained to classify the digit ``zero'' in the Digits dataset. Each subplot shows the weight patterns for individual connection patterns, labeled according to connection types: P-P (positive input to positive hidden sublayers), N-P (negative input to positive hidden sublayers), P-N (positive input to negative hidden sublayers), and N-N (negative input to negative hidden sublayers). Color gradients denote weight values, with yellow indicating strongly positive weights and purple indicating strongly negative weights. The seed for this experiment is 0.
}
 \label{fig:edla_hidden_weights}
\end{figure}

Figure \ref{fig:edla_hidden_weights} shows a heatmap of the synaptic weights connecting the input layer to the hidden layer within the EDLA network, specifically trained to classify the digit ``zero'' from handwritten digits. The heatmap reveals that the network learned meaningful features from the input data. Notably, the neurons in the hidden layer exhibit a strong response to specific patterns within the images. For instance, the weights connecting the positive neurons in the input layer to the positive neurons in the hidden layer (denoted as "P-P" connections) show high activation, shaping the digit ``zero.'' This result suggests that the EDLA method effectively enables neurons to capture and represent the relevant digit-specific features. Additionally, similar ``zero-shaped'' patterns also emerge in other synaptic connections.

\begin{figure}
 \centering
 \includegraphics[width=0.8\linewidth]{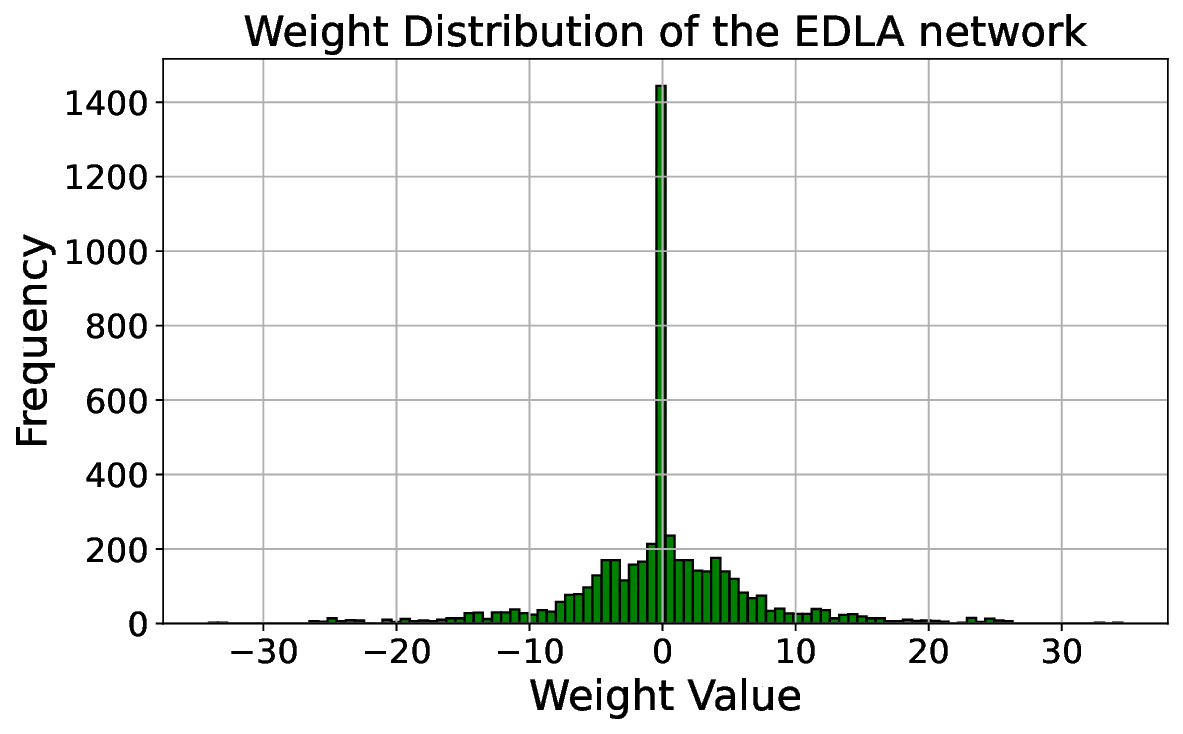}
 \caption{Histogram of synaptic weight distribution in the EDLA Network. The horizontal axis denotes the synaptic weight values, and the vertical axis indicates the corresponding frequencies. The seed for this experiment is 0.}
 \label{fig:edla_distribution}
\end{figure}

Synaptic weight distributions in neural networks provide valuable insights into the learning behavior of the network and features extracted from the input data. To explore this further, we analyze the distribution of weights across the EDLA network. Figure \ref{fig:edla_distribution} shows a histogram representing the synaptic weights within the trained EDLA network. The distribution appears to be predominantly centered around small magnitudes, indicating that most weights are relatively small. However, a distinct spread is observed with a smaller subset of synapses exhibiting large positive or negative values. This broad distribution suggests that the EDLA update unilaterally increases the absolute values of the weights while balancing the excitatory and inhibitory weights.

\subsection{Stabilization of Learning Dynamics in EDLA with ReLU Activation Function}
\label{sec:stabilization}

Deeper EDLA networks employing the ReLU AF often exhibit severe performance degradation when the number of hidden layers (depth) increases. This degradation arises from the unbounded nature of the ReLU AF, which can generate excessively large activations and thereby produce excessively large and unstable weight updates during training. To mitigate these effects, two stabilization techniques are introduced: the use of small-scale initial weights and RMSNorm of neuronal activations (Sec. \ref{sec:rms}). The former limits the magnitude of initial activations, while the latter normalizes neuronal pre-activations based on their root-mean-square (RMS) values. These mechanisms jointly stabilize the learning dynamics by suppressing excessive activations and preventing abrupt weight updates.

\begin{figure}
 \begin{center}
 \includegraphics[width=1\linewidth]{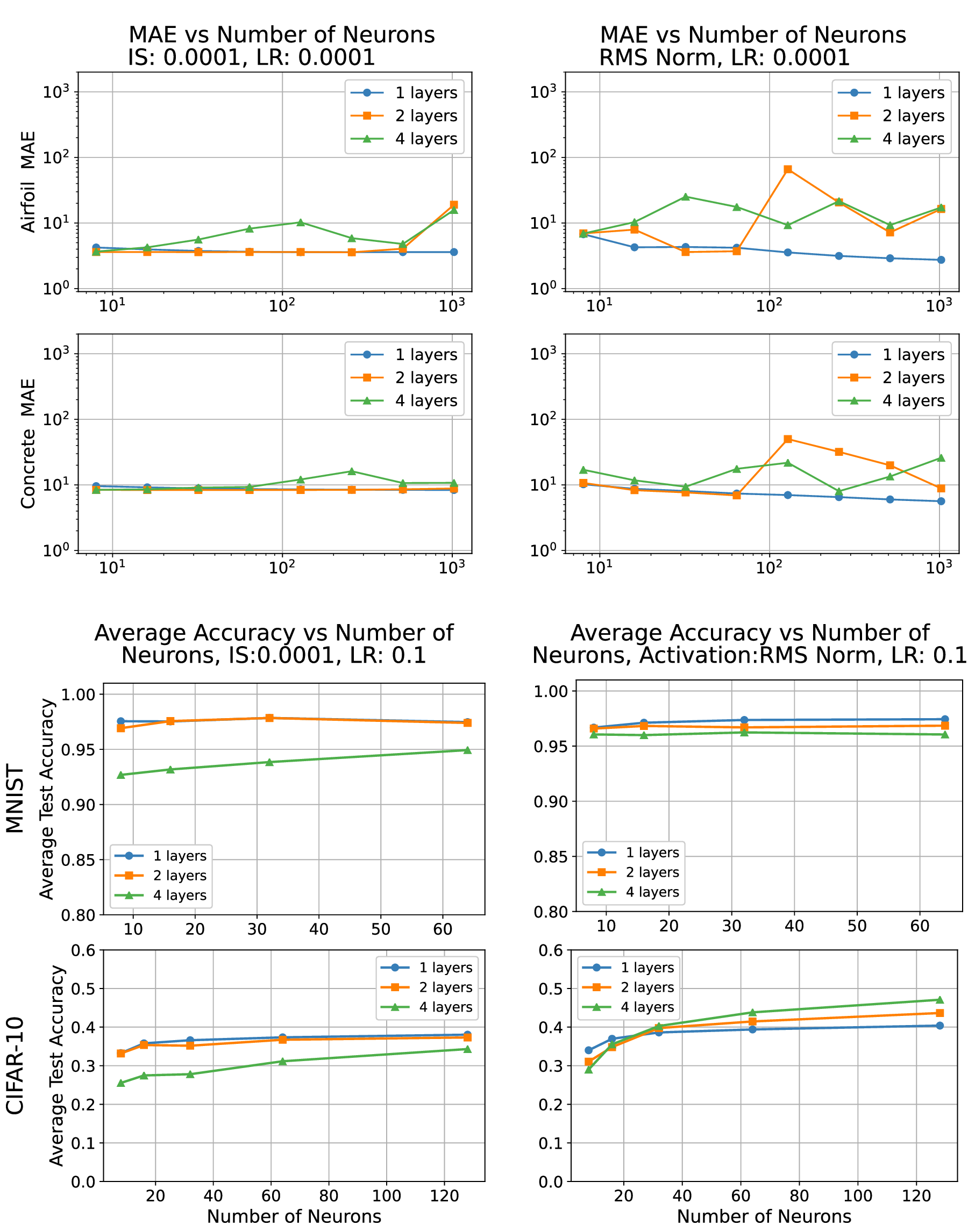}
 \caption{Performance comparison of EDLA with ReLU AF using two stabilization techniques: small initial weights (IS = 0.0001, left column) and RMSNorm (right column). The plots show performance across varying layer configurations (1, 2, and 4 layers) and neuron counts. Top two rows (Regression): MAE on Airfoil and Concrete datasets (LR = 0.0001). Bottom two rows (Classification): Accuracy on MNIST and CIFAR-10 datasets (LR = 0.1). Results are averaged over four independent trials. The seeds for regression tasks are the same as those used in Fig.~\ref{fig:edla_regression}, and the seeds for classification tasks are the same as those used in Fig.~\ref{fig:edla_sigmoid} to ensure consistency.}
 \label{fig:edla_improvement}
 \end{center}
\end{figure}

Figure \ref{fig:edla_improvement} presents the results of EDLA networks with the ReLU AF under these stabilization strategies across a range of network widths and depths for both regression and image classification tasks. For regression (Airfoil and Concrete datasets) and classification (MNIST and CIFAR-10 datasets), the learning rates are set to 0.0001 and 0.1, respectively. The weights are drawn from uniform distributions over $[0, \mathrm{IS}]$ and $[-\mathrm{IS}, 0]$ for synapses connecting same-type and different-type neurons, respectively. The initial scale (IS) was set to 0.0001 for the small-scale initialization variant, whereas the standard baseline and the RMSNorm variants used a default scale of 1.0. RMSNorm is applied to each layer except the input and output layers. The mini-batch sizes are 64 for regression and 128 for classification.

For regression tasks, both small initial weights and RMSNorm suppress extreme or NaN (not-a-number) MAE values that arise in deeper networks (see Fig.~\ref{fig:edla_regression}), leading to stable training. For classification tasks, while small initial weights do not significantly alter the accuracy of 1-layer networks, they can improve the performance of 2-layer configurations on the CIFAR-10 dataset, elevating their accuracy to match that of the 1-layer configuration. In deeper configurations (4 layers) on MNIST, they tend to degrade accuracy slightly. In contrast, RMSNorm improves accuracy in deeper architectures, particularly for the CIFAR-10 dataset with larger hidden dimensions. These findings indicate that small initial weights and RMSNorm are effective in preventing instability in deeper EDLA networks with the ReLU AF, stabilizing training without compromising general performance.

The observed improvement in deeper EDLA networks is consistent with stabilization techniques suppressing extreme internal dynamics. The instability of ReLU-based EDLA networks can be explained by two interacting factors: (i) large activation outliers and (ii) the dying ReLU phenomenon, in which neurons become inactive and cease to contribute to learning. The former refers to the uncontrolled growth of activations during forward propagation, which induces large Hebbian-like weight updates under EDLA's learning rule. This effect becomes increasingly pronounced as the depth of the network grows, resembling the well-known exploding gradient phenomenon in recurrent or very deep networks \citep{He:2015,Pascanu:2013}. The latter, known as the dying ReLU problem \citep{Douglas:2018,Lu:2020}, occurs when neurons become inactive (outputting zero for all inputs), preventing them from contributing to learning. In practice, these two effects interact: large activations in early training produce extreme weight updates, which then push many neurons into permanently inactive states, further deteriorating the network's representational capacity.

\begin{figure}
 \begin{center}
 \includegraphics[width=1\linewidth]{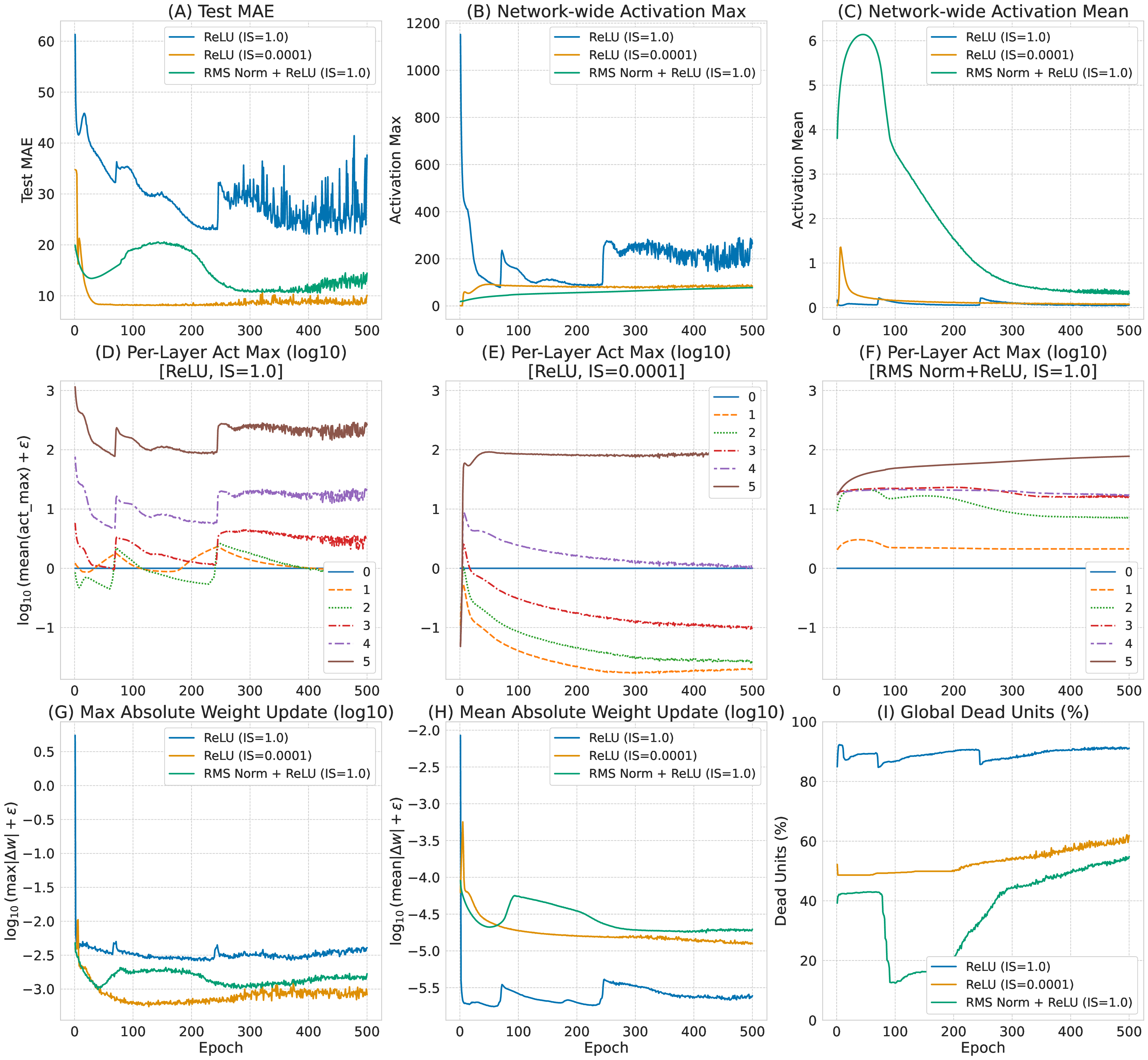}
 \caption{Analysis of internal dynamics for three EDLA-ReLU variants on the Concrete regression task ($n_\mathrm{hidden} = 32$, $L=4$, LR = 0.0001). Variants are: ReLU (scale = 1.0) (baseline EDLA), ReLU (scale = 0.0001) (EDLA with the small initial scale), and RMS Norm + ReLU (scale = 1.0) (EDLA with RMSNorm). Metrics shown: (A) Test MAE; (B) network-wide maximum activation; (C) network-wide mean activation; (D--F) per-layer post-activation maxima (layer index 0: input, 1--4: hidden layers, 5: output); (G) network-wide maximum absolute weight update; (H) network-wide mean absolute weight update; (I) global dead-unit fraction. Metric (A) was evaluated on the test set, while metrics (B)--(I) were evaluated on the training set. Activation statistics in (B--F) are computed on post-activations (after ReLU; after RMSNorm when enabled). (D--H) are plotted on a log10 scale ($\varepsilon = 10^{-12}$ added to avoid $\log(0)$).  The results are averaged over four independent trials using specific random seeds (5698, 57443, 96581, 97484).}
\label{fig:analysis_relu_stabilization}
 \end{center}
\end{figure}

To elucidate the effects of these stabilization methods and provide quantitative diagnostics supporting our stability claims, several internal metrics were monitored. Test MAE was evaluated on the test set to track generalization performance. Concurrently, activation statistics and update magnitudes were monitored on the training data, including the network-wide maximum and mean activations, per-layer activation maxima, the network-wide maximum and mean absolute weight updates, and the global dead-unit ratio (the percentage of inactive neurons). In this study, a neuron was considered dead if its activation remained below the fixed threshold of 0.01 for more than 95\% of training samples. Figure \ref{fig:analysis_relu_stabilization} compares these metrics across three EDLA variants: the baseline EDLA with the ReLU AF, the EDLA with small-scale initialization (IS = 0.0001), and the EDLA with RMSNorm. The initial scale (IS) was set to 1.0 for both the baseline EDLA and the EDLA with RMSNorm. This internal analysis was conducted on the Concrete regression task using a network with four hidden layers of 32 neurons each, a learning rate of 0.0001, and a mini-batch size of 64.

As shown in Figure \ref{fig:analysis_relu_stabilization}A, the baseline EDLA exhibits large and unstable test MAE values throughout training, indicating poor convergence and sensitivity to parameter updates. In contrast, both small initial weights and RMSNorm suppress large fluctuations in MAE, leading to smoother and smaller test errors. The small-scale initialization achieves the lowest MAE overall, while RMSNorm provides similarly stable and improved performance. Figures \ref{fig:analysis_relu_stabilization}B and \ref{fig:analysis_relu_stabilization}C show that the baseline network experiences a rapid increase in the maximum activation magnitude early in training. The small-scale initialization moderates this activation growth, and RMSNorm stabilizes activations at moderate levels throughout training. Interestingly, the RMS-normalized network maintains higher mean activations than the other variants, even when its maximum activations are well controlled, suggesting that RMSNorm reduces neuron inactivity while maintaining balanced activation levels. This interpretation is supported by the lower dead-unit ratio observed in Figure \ref{fig:analysis_relu_stabilization}I.

The per-layer activation trends in Figures \ref{fig:analysis_relu_stabilization}D--F reveal that deeper layers tend to exhibit larger activations due to the cumulative effect of the unbounded ReLU activation, while both stabilization techniques attenuate this amplification across the hidden layers. Figures \ref{fig:analysis_relu_stabilization}G and \ref{fig:analysis_relu_stabilization}H further track the update magnitudes, showing that the baseline EDLA undergoes excessively large weight updates in the early stages of training. In the standard EDLA, although the mean weight update is smaller than that of the stabilized variants, the weights of many neurons remain nearly unchanged, while a few experience extremely large updates (i.e., outlier-dominated / heavy-tailed updates), leading to unstable learning dynamics. Consequently, the global dead-unit fraction shown in Figure \ref{fig:analysis_relu_stabilization}I reveals that the baseline model suffers from a high proportion of inactive neurons. Many neurons become permanently dead early in training, severely limiting the network's representational capacity. In contrast, both stabilization methods maintain smaller and more consistent weight update magnitudes across epochs. Both stabilization methods substantially reduce the occurrence of dead units, although a gradual increase is observed in the later stages of training. These results indicate that the stabilization techniques effectively regulate activation and weight dynamics.

These results collectively demonstrate that the instability of ReLU-based EDLA networks arises from the combined effects of exploding activations and dead ReLU units. Small initial weights alleviate early activation explosions, whereas RMSNorm further stabilizes learning by controlling both activation magnitudes and update scales. Importantly, RMSNorm maintains sufficient neuron activity and prevents the collapse of representational diversity. Overall, these findings indicate that appropriate control of activation magnitudes is essential for stable training in EDLA networks. The effectiveness of RMSNorm in particular parallels the success of normalization techniques such as Batch Normalization \citep{Ioffe:2015} and Layer Normalization \citep{Ba:2016}, highlighting that regulation of internal activation dynamics is a key factor for achieving stable and efficient learning in non-standard architectures such as EDLA.

\subsection{Comparison with multilayer perceptron trained with backpropagation}

To benchmark the performance of the EDLA network, it is compared with a standard multilayer perceptron (MLP) trained using backpropagation. The MLP baseline follows the same depth regime as the EDLA setting (input layer, $L$ hidden layers, and an output layer), and we choose widths from the same candidate set (or the closest available choice) to obtain comparable model sizes. Hidden units use either sigmoid or ReLU activations. For regression, the output layer is linear and we minimize mean squared error (MSE). For classification, the output layer produces logits (linear outputs) and we minimize the softmax cross-entropy loss. The softmax cross-entropy loss is employed because it is the standard loss function for multi-class classification tasks.

We train the MLP using AdamW (weight decay = 0; other hyperparameters at PyTorch defaults) with a constant learning rate (no scheduling). We tune only the learning rate via a small sweep for the Concrete dataset (one hidden layer, 1024 hidden neurons) and the MNIST dataset (one hidden layer, 512 hidden neurons). Specifically, for regression we evaluate $\eta \in \{10^{-3}, 10^{-2}, 10^{-1}\}$ for MLP-Sigmoid and $\eta \in \{10^{-4}, 10^{-3}, 10^{-2}\}$ for MLP-ReLU; for classification we evaluate $\eta \in \{10^{-3}, 10^{-4}\}$ for MLP-Sigmoid and $\eta \in \{10^{-4}, 10^{-5}\}$ for MLP-ReLU. We select the learning rate based on the best test performance (MAE for regression; accuracy for classification). This procedure yields $\eta=0.1$ (Sigmoid) and $\eta=0.01$ (ReLU) for regression, and $\eta=10^{-3}$ (Sigmoid) and $\eta=10^{-4}$ (ReLU) for classification. Training runs for 500 epochs with batch size 64 for regression and 128 for classification (1000 epochs for CIFAR-10). All EDLA training configurations follow the corresponding settings described in the previous sections. 

While stabilization techniques such as RMSNorm were necessary to prevent catastrophic failures in deep EDLA-ReLU networks, we do not apply RMSNorm to the baseline MLPs because BP-trained MLPs are already stable under the same depth/width regimes. We include a standard BP-trained MLP as a reference upper bound to contextualize EDLA's achievable performance on each task. Our goal is not to claim superiority of EDLA over BP, but to clarify the performance gap and characterize EDLA's limitations. Accordingly, BP serves as a reference model rather than a component-matched competitor, and we keep it in a standard training configuration (AdamW, constant learning rate, no scheduling, no regularization unless otherwise stated). Consequently, we do not attempt to match EDLA-specific stabilizers in the BP baselines, as doing so is not required for our purpose.

Table \ref{tab:performance_selected} presents the performance comparison across four representative datasets. In the ``Model (Act)'' column, the activation function and stabilization variant are indicated in parentheses. ``IS'' refers to the initial scale parameter used for weight initialization, and ``RMS'' indicates the application of RMSNorm to the pre-activations. For example, ``EDLA (ReLU, IS=0.0001)'' represents the ReLU-based EDLA with a small initial scale of 0.0001, and ``EDLA (ReLU+RMS)'' represents the ReLU-based EDLA incorporating RMSNorm.

For a fair comparison, the parameter counts of EDLA and MLP in each configuration were matched or approximated as closely as possible. Across all datasets, the best BP-trained MLP outperforms the best EDLA variant under our matched/near-matched settings, while the gap depends on task complexity and depth. EDLA can reach high accuracy in several shallow settings, although the best EDLA variant remains below the best BP baseline under our matched/near-matched settings. On the 1-layer MNIST task, for instance, the best EDLA variant (ReLU, IS=0.0001) achieves 0.975 accuracy, only slightly trailing the MLP's 0.982. In contrast, the gap widens significantly on deeper, more complex tasks. On the 4-layer Airfoil regression task, the MLP (ReLU) achieves an MAE of 1.868, whereas the best-performing stabilized EDLA (ReLU, IS=0.0001) only reaches 5.873. This highlights a key finding: while the stabilization techniques (Sec. \ref{sec:stabilization}) successfully prevented the catastrophic failure of deep EDLA-ReLU networks (which produced MAE $> 10^{30}$), they were not sufficient to close the fundamental optimization gap with a standard backpropagation-trained MLP. For multi-class problems, EDLA necessarily uses $K$ independent single-output networks; therefore, exact parameter matching to a single multi-output MLP is generally not possible, and we report parameter counts explicitly for transparency.

This gap is consistent with EDLA's use of a single diffused error signal rather than layer-specific gradients. Backpropagation achieves its high performance by propagating exact, layer-specific error gradients, allowing for fine-grained optimization across the entire network. EDLA, by design, sacrifices this optimization power for biological plausibility, relying instead on an approximated, single global error signal diffused uniformly to all layers. Furthermore, this gap is compounded on multi-class tasks (MNIST, CIFAR-10) by EDLA's parallel architecture. An MLP uses a shared representational body and a single final layer for all classes. In contrast, the EDLA implementation must train $K$ independent networks, one for each output class. This approach not only requires significantly more parameters (as analyzed in Sec. \ref{sec:param_count}) but also prevents the network from learning shared, general-purpose features across classes, likely degrading performance on high-dimensional, multi-class problems.

On the CIFAR-10 dataset, the MLP's performance is also relatively low, but this result is not irregular for a simple MLP architecture. Simple MLPs with ReLU show about 50\% accuracy on CIFAR-10. For example, a simple MLP with ReLU activation, six layers, and 1024 neurons per layer achieves 54.2\% accuracy with Layer Normalization on CIFAR-10 \citep{Bachmann:2023}, while a two-layer MLP (width 512, ReLU, and 0.2 dropout) achieves 54--58\% accuracy \citep{Mukkamala:2017}. Thus, the MLP's performance in the experiments is consistent with the findings in the literature, confirming that both EDLA and traditional MLPs encounter similar architectural limitations when applied to high-dimensional image data without convolutional priors.

\begin{table}[htbp]
\centering
\footnotesize
\caption{Performance comparison on MNIST, CIFAR-10, Airfoil, and Concrete. For classification the metric is accuracy (higher is better); for regression the metric is mean absolute error (MAE; lower is better). For each dataset, the best value in the table is shown in bold. Ties are also bolded. Parameter counts for both EDLA and MLP include bias terms. All values are reported as mean $\pm$ SD over four independent runs. For regression, we use the same seeds as in Fig.~\ref{fig:edla_regression}; for classification, we use the same seeds as in Fig.~\ref{fig:edla_sigmoid}.}
\label{tab:performance_selected}
\begin{tabular}{llllll}
\toprule
Dataset & Metric & Model (Act) & Arch (L x N) & Params & Value (Mean $\pm$ SD) \\
\midrule
 Airfoil & MAE   & EDLA (Sigmoid)           &  1 x 256   &  7.2k  &   $3.560               \pm 0.105$    \\
         & MAE   &   EDLA (ReLU)            &  1 x 256   &  7.2k  &   $3.323               \pm 0.091$    \\
         & MAE   &  EDLA (ReLU, IS=0.0001)  &  1 x 256   &  7.2k  &   $3.592               \pm 0.069$    \\
         & MAE   & EDLA (ReLU+RMS)          &  1 x 256   &  7.2k  &   $3.147               \pm 0.041$    \\
         & MAE   &  \textbf{MLP (Sigmoid)}  &  1 x 1024  &  7.2k  &   $\mathbf{1.379}      \pm 0.049$         \\
         & MAE   &   MLP (ReLU)             &  1 x 1024  &  7.2k  &   $1.673               \pm 0.065$    \\
         & MAE   & EDLA (Sigmoid)           &  4 x 256   & 1.58M  &   $5.922               \pm 1.106$    \\
         & MAE   &   EDLA (ReLU)            &  4 x 256   & 1.58M  &  $>10^{30}$ (diverged)  \\
         & MAE   &   EDLA (ReLU, IS=0.0001) &  4 x 256   & 1.58M  &  $ 5.873               \pm 2.697  $\\
         & MAE   & EDLA (ReLU+RMS)          &  4 x 256   & 1.58M  &  $21.533               \pm 22.246 $\\
         & MAE   &  MLP (Sigmoid)           &  4 x 768   & 1.78M  &  $ 5.788               \pm 0.144  $\\
         & MAE   &   MLP (ReLU)             &  4 x 768   & 1.78M  &  $ 1.868               \pm 0.108  $\\
\midrule
Concrete & MAE  &  EDLA (Sigmoid)           &  1 x 256   & 10.2k  &   $7.833               \pm 0.258$  \\
         & MAE  &    EDLA (ReLU)            &  1 x 256   & 10.2k  &   $7.430               \pm 0.292$  \\
         & MAE  &   EDLA (ReLU, IS=0.0001)  &  1 x 256   & 10.2k  &   $8.449               \pm 0.397$  \\
         & MAE  &  EDLA (ReLU+RMS)          &  1 x 256   & 10.2k  &   $6.513               \pm 0.246$  \\
         & MAE  &   MLP (Sigmoid)           &  1 x 1024  & 10.2k  &   $3.634               \pm 0.331$  \\
         & MAE  &    MLP (ReLU)             &  1 x 1024  & 10.2k  &   $3.539               \pm 0.352$  \\
         & MAE  &  EDLA (Sigmoid)           &  4 x 256   & 1.59M  &   $9.033               \pm 1.542$  \\
         & MAE  &    EDLA (ReLU)            &  4 x 256   & 1.59M  &  $>10^{30}$ (diverged)     \\
         & MAE  &   EDLA (ReLU, IS=0.0001)  &  4 x 256   & 1.59M  &   $16.165              \pm 4.709$ \\
         & MAE  &  EDLA (ReLU+RMS)          &  4 x 256   & 1.59M  &   $8.017               \pm 0.877$ \\
         & MAE  &   MLP (Sigmoid)           &  4 x 768   & 1.78M  &   $13.925              \pm 0.206$ \\
         & MAE  &  \textbf{MLP (ReLU)}     &  4 x 768   & 1.78M   &   $\mathbf{3.191}      \pm 0.240$ \\
\midrule
   MNIST & Accuracy &EDLA (Sigmoid)           &   1 x 16   & 503.1k&    $0.972           \pm 0.001$\\
         & Accuracy &  EDLA (ReLU)            &   1 x 16   & 503.1k&    $0.969           \pm 0.001$\\
         & Accuracy &EDLA (ReLU, IS=0.0001)   &   1 x 16   & 503.1k&    $0.975           \pm 0.003$\\
         & Accuracy &EDLA (ReLU+RMS)          &   1 x 16   & 503.1k&    $0.971           \pm 0.001$\\
         & Accuracy &MLP (Sigmoid)            &  1 x 512   & 407.1k&    $0.982           \pm 0.001$\\
         & Accuracy &MLP (ReLU)               &  1 x 512   & 407.1k&    $0.981           \pm 0.000$\\
         & Accuracy &EDLA (Sigmoid)           &   4 x 64   & 3.00M &    $0.968           \pm 0.001$\\
         & Accuracy &  EDLA (ReLU)            &   4 x 64   & 3.00M &    $0.954           \pm 0.007$\\
         & Accuracy &EDLA (ReLU, IS=0.0001)   &   4 x 64   & 3.00M &    $0.949           \pm 0.002$\\
         & Accuracy &EDLA (ReLU+RMS)          &   4 x 64   & 3.00M &    $0.961           \pm 0.001$\\
         & Accuracy &\textbf{MLP (Sigmoid)}   &  4 x 1024  & 3.96M &    $\mathbf{0.985}  \pm 0.000$\\
         & Accuracy & \textbf{MLP (ReLU)}     &  4 x 1024  & 3.96M &    $\mathbf{0.985}  \pm 0.000$\\
\midrule
CIFAR-10 & Accuracy &EDLA (Sigmoid)           &   1 x 32   & 3.93M &    $0.417          \pm 0.004$\\
         & Accuracy &  EDLA (ReLU)            &   1 x 32   & 3.93M &    $0.409          \pm 0.003$\\
         & Accuracy & EDLA (ReLU, IS=0.0001)  &   1 x 32   & 3.93M &    $0.366          \pm 0.002$\\
         & Accuracy &EDLA (ReLU+RMS)          &   1 x 32   & 3.93M &    $0.386          \pm 0.002$\\
         & Accuracy &MLP (Sigmoid)            &  1 x 1024  & 3.16M &    $0.467          \pm 0.001$\\
         & Accuracy & MLP (ReLU)              &  1 x 1024  & 3.16M &    $0.544          \pm 0.002$\\
         & Accuracy &EDLA (Sigmoid)           &   4 x 16   & 2.03M &    $0.138          \pm 0.005$\\
         & Accuracy &  EDLA (ReLU)            &   4 x 16   & 2.03M &    $0.233          \pm 0.028$\\
         & Accuracy &EDLA (ReLU, IS=0.0001)   &   4 x 16   & 2.03M &    $0.275          \pm 0.024$\\
         & Accuracy &EDLA (ReLU+RMS)          &   4 x 16   & 2.03M &    $0.355          \pm 0.012$\\
         & Accuracy &MLP (Sigmoid)            &  4 x 512   & 2.37M &    $0.474          \pm 0.003$\\
         & Accuracy & \textbf{MLP (ReLU)}     &  4 x 512   & 2.37M &    $\mathbf{0.552} \pm 0.004$\\
\bottomrule
\end{tabular}
\end{table}

\section{Conclusion and Discussion}

The Error Diffusion Learning Algorithm (EDLA) provides a biologically inspired alternative to conventional backpropagation methods for training neural networks and is characterized by the diffusion of a global error signal across network layers. This mechanism is analogous to reward-based learning processes observed in biological neural systems. Moreover, EDLA mirrors neuromodulatory systems, such as the dopaminergic system, which projects broadly across cortical and subcortical areas, disseminating their signals globally \citep{Roelfsema:2018}. In contrast to traditional MLPs, EDLA incorporates structured interactions between neurons with excitatory and inhibitory synapses. This reflects the biologically observed balance between these types of synapses.

However, EDLA is inherently constrained by its single-output architecture, limiting its applicability to multiclass classification tasks. A practical solution involves parallelizing multiple EDLA networks, although this approach increases computational costs. This parallelized architecture mirrors the modular structure of biological neural systems in which specialized circuits process distinct features or tasks independently. In such a configuration, each EDLA network separately optimizes its own output according to a unique global error signal to that network. This modular structure aligns with distributed processing mechanisms observed in biological neural networks.

Our experimental evaluation shows that EDLA can achieve satisfactory performance across parity-check, regression, and image classification tasks, particularly in shallow configurations. Notably, networks using the ReLU activation function often achieved final performance comparable to those using the sigmoid function originally proposed by Kaneko \citep{Kaneko}. Furthermore, ReLU-based EDLA can converge faster in some controlled settings (e.g., parity). This trend is consistent with earlier observations that ReLU activations can accelerate learning dynamics in standard neural networks \citep{Nair:2010}. However, this advantage is not consistent across benchmarks. At the same time, our results highlight an important sensitivity: with unbounded activations of ReLU, EDLA can become more sensitive to step size and depth.

A central finding of this study is that increasing network depth in EDLA does not reliably improve performance and can lead to degradation, unlike in standard deep learning (Figs.~\ref{fig:edla_regression}, \ref{fig:edla_sigmoid}, \ref{fig:edla_relu}). We hypothesize this stems from EDLA's unique update mechanism based on a single global error signal. Unlike backpropagation, where gradients can vanish in deep layers due to the chain product of activation function derivatives \citep{Bengio:1994,Pascanu:2013}, EDLA's global signal ensures all weights receive an update signal. While this avoids the classical vanishing gradient problem caused by the chain product of derivatives, it does not guarantee effective credit assignment for deep hierarchies. Updates driven by the single scalar error signal $d$ might not be optimal for each specific layer's needs. Furthermore, changes induced in lower-layer activations propagate upwards and can be amplified through successive layers. This amplification becomes particularly severe (``exploding activations'') when using unbounded activation functions like ReLU. However, even with bounded functions like sigmoid, the cumulative effect of layer-by-layer amplification driven by a non-specific global signal might interfere with learning the complex, hierarchical features typically expected from deeper networks, potentially explaining the lack of performance improvement or degradation. This amplification effect is inherently less pronounced in shallower networks. Consequently, for EDLA, increasing network width appears to be a more effective and stable strategy for improving performance than increasing depth.

We also identified a depth-related instability observed, especially with the ReLU activation function. This instability manifested severely in deep ReLU regression networks, causing catastrophic failures (e.g., NaN MAE). This failure mode is likely exacerbated by the regression task's linear output layer: exploding activations from the final hidden layer directly produce an extremely large output $y$, which in turn generates an enormous global error signal $d = t - y$. This huge error signal then catastrophically amplifies weight updates throughout the entire network. This depth-related instability primarily stems from the unbounded nature of the ReLU AF itself, a phenomenon observed in standard deep networks as well \citep{He:2015, Pascanu:2013}. As signals propagate through multiple layers, ReLU's lack of saturation allows activation magnitudes to grow excessively (``exploding activations''). While this is a general challenge in deep ReLU networks, it becomes particularly problematic within the EDLA framework. EDLA's Hebbian-like learning rule translates these large activations directly into excessively large weight updates. The single global error signal, while perhaps not the primary initiator of the explosion, likely contributes to propagating instability across the network once large activations emerge.

Our internal dynamics analysis (Fig.~\ref{fig:analysis_relu_stabilization}) provides quantitative evidence consistent with this cascade and reveals a mechanism for vanishing updates distinct from classical gradient vanishing. The baseline deep ReLU network exhibited exploding maximum activations (Fig.~\ref{fig:analysis_relu_stabilization}B) and extreme early weight updates (Fig.~\ref{fig:analysis_relu_stabilization}G,H), followed by a rapid rise in the dead-unit fraction to over 80\% (Fig.~\ref{fig:analysis_relu_stabilization}I). This suggests that performance collapse stems from ReLU-driven exploding activations which, exacerbated by EDLA's update rule, trigger widespread neuron death. These dead units then provide near-zero update signals, effectively halting learning. In contrast, the saturating sigmoid function inherently limits activation magnitudes and update sizes, contributing to greater stability in this scheme.

It was demonstrated that this instability can be effectively managed. While smaller learning rates and initial weights help, RMSNorm proved particularly robust, controlling extreme activations while crucially maintaining neuron activity (low dead unit fraction, Fig. \ref{fig:analysis_relu_stabilization}I), thus enabling stable learning in deeper ReLU networks.

Despite successful stabilization, comparison with backpropagation-trained MLPs (Table \ref{tab:performance_selected}) revealed a persistent performance gap, widening with network depth and task complexity. Although EDLA achieved highly competitive results in shallow configurations, MLPs consistently performed better overall. This underscores EDLA's fundamental trade-off: sacrificing the optimization power of exact, layer-specific gradients for the biological plausibility of an approximated, global error signal. The parallel architecture likely exacerbates this gap in multi-class settings by preventing shared feature learning.

Overall, rather than positioning EDLA as a direct competitor to backpropagation, our results provide a diagnostic characterization of where and why global error diffusion breaks down in contemporary deep-network settings, and establish a concrete baseline for future work on local, biologically inspired learning rules.

EDLA employs a form of \emph{structural modularity}, in which computational units are explicitly separated by design. We interpret this as analogous to evolutionarily and developmentally shaped anatomical specialization in biological systems, which can yield partially segregated processing subsystems over long timescales. In modern machine learning, related design choices underpin successful large-scale systems: for example, Mixture-of-Experts (MoE) architectures decompose computation into expert subnetworks controlled by gating mechanisms, and ensemble methods exploit multiple models to improve robustness and predictive uncertainty \citep{Fedus:2022,Lakshminarayanan:2017}. Biology likewise contains specialized subsystems that instantiate structural modularity as a recurring circuit motif, including parallel information-processing channels originating in the retina \citep{daSilveira:2011}, and cerebellar microzone organization with instructive climbing-fiber signals \citep{Apps:2018}.

However, the modular organization observed in the neocortex, such as barrel cortex maps discussed in Sec.~\ref{sec:multi_output}, is not purely innate, but can also emerge through activity- and experience-dependent map formation \citep{Inan:2007}. This motivates \emph{learned modularization} as an important direction for future EDLA variants. Driving robust self-organized module formation will likely require additional biologically grounded inductive biases, particularly \emph{locality} constraints. One promising avenue is the introduction of spatial embedding and wiring-economy constraints. Just as evolutionary simulations suggest that selection to minimize connection costs can promote the evolution of modularity over biological timescales \citep{Clune:2013}, enforcing analogous economy constraints during network training inherently encourages distance-dependent connectivity and effective locality of interactions, thereby promoting spatial clustering and small-world organization \citep{Achterberg:2022}. A second promising avenue is spatially constrained neuromodulatory volume transmission \citep{Ozete:2024}. The latter is particularly relevant as a working hypothesis for extending EDLA's diffusion-like teaching signal: localizing the currently global error signal in space may support context factorization and reduce interference, providing a plausible route from global error diffusion toward biologically plausible local plasticity \citep{Bull:2025}.

Future work could explore several avenues to enhance EDLA's capabilities. First, integrating eligibility traces could enable single EDLA networks to handle multiclass classification tasks, reducing the need for parallel architectures. Eligibility traces could allow neurons to maintain a memory of past activations, facilitating more complex decision boundaries. However, determining which neurons should maintain these traces remains an open question. Second, investigating hybrid architectures that combine EDLA with backpropagation in a complementary manner might leverage the strengths of both approaches. For instance, EDLA might be employed for initial coarse learning phases, followed by fine-tuning via backpropagation. Finally, applying EDLA to other architectures such as recurrent and convolutional neural networks \citep{Fukushima:1980,Krizhevsky:2012,LeCun:1998} could extend its applicability to temporal and spatial data domains, respectively.

\section*{Acknowledgments}

The author gratefully acknowledges the assistance of Chat AI assistants (OpenAI ChatGPT and Google Gemini) and grammar checking service (Grammarly) in improving the grammar and style of this work.

\section*{Declarations}

\subsection*{Preprint}
A preliminary version of this manuscript was posted as a preprint (arXiv:2504.14814).

\subsection*{Data Availability}

The source code generated during this study is publicly available on GitHub at \url{https://github.com/KazuhisaFujita/EDLA}. The datasets analyzed during this study are publicly available in the following repositories:
\begin{itemize}
    \item Airfoil Self-Noise dataset, UCI Machine Learning Repository [DOI: 10.24432/C5VW2C]:  
    \url{https://archive.ics.uci.edu/dataset/291/airfoil+self+noise}
    \item Concrete Compressive Strength dataset, UCI Machine Learning Repository [DOI: 10.24432/C50P4X]:  
    \url{https://archive.ics.uci.edu/ml/datasets/Concrete+Compressive+Strength}
    \item Energy Efficiency dataset, UCI Machine Learning Repository [DOI: 10.24432/C5PK67]:  
    \url{https://archive.ics.uci.edu/ml/datasets/Energy+efficiency}
    \item Optical Recognition of Handwritten Digits dataset, UCI Machine Learning Repository [DOI: 10.24432/C50P49]:  
    \url{https://archive.ics.uci.edu/dataset/80/optical+recognition+of+handwritten+digits}
    \item MNIST dataset: \url{http://yann.lecun.com/exdb/mnist/}  
    \item Fashion-MNIST dataset: \url{https://github.com/zalandoresearch/fashion-mnist}
    \item CIFAR-10 dataset: \url{https://www.cs.toronto.edu/~kriz/cifar.html}
\end{itemize}

\subsection*{Code availability}
The source code used in this study is available at \url{https://github.com/KazuhisaFujita/EDLA}.

\subsection*{Funding Declaration}

The author received no funding for this work.

\subsubsection*{Conflict of Interest}

The author declares no competing interests.

\subsubsection*{Ethical Approval}

This article does not contain any studies with human participants or animals performed by the author.


\appendix
\section*{Appendix (for pedagogical completeness)}
\section{Derivation of Backpropagation in a Multilayer Perceptron}

A multilayer perceptron (MLP) is a class of feedforward artificial neural networks characterized by multiple interconnected neuron layers: an input layer, one or more hidden layers, and an output layer. Typically, each neuron within one layer is fully connected to all neurons in the subsequent layer.

The input to an MLP is represented by the vector $\mathbf{x} = [x_0, x_1, \ldots, x_N]^\mathrm{T} \in \mathbb{R}^{N+1}$, where $x_0$ commonly serves as a bias term (usually set as $x_0 = 1$) and $N$ denotes the number of input features. This input vector is received by the input layer and subsequently propagated forward to the first hidden layer.

For neuron $j$ in the first hidden layer $(l=1)$, the activation $a_j^{(1)}$ is calculated as the weighted sum of its inputs:
\begin{equation}
a_j^{(1)} = \mathbf{w}_j^{(1)\mathrm{T}} \mathbf{x}
\end{equation}
where $\mathbf{w}_j^{(1)} = [w_{j0}^{(1)}, w_{j1}^{(1)}, \ldots, w_{jN}^{(1)}]^\mathrm{T}$ represents the weight vector connecting the input layer to neuron $j$ in the first hidden layer $(l=1)$. Here, $w_{ji}^{(1)}$ denotes the weight connecting from the neuron $i$ in the input layer to the neuron $j$ in the first hidden layer. The output of neuron $j$ is then obtained by applying the activation function $g(\cdot)$:
\begin{equation}
z_j^{(1)} = g(a_j^{(1)}).
\end{equation}
For subsequent hidden layers $(l = 2, ..., L)$, the activation $a_j^{(l)}$ for neuron $j$ in layer $l$ is computed as follows:
\begin{equation}
a_j^{(l)} = \mathbf{w}_j^{(l)\mathrm{T}} \mathbf{z}^{(l-1)},
\end{equation}
where $\mathbf{z}^{(l)} = [z_0^{(l)}, z_1^{(l)}, \ldots, z_{M^{(l)}}^{(l)}]^\mathrm{T} \in \mathbb{R}^{M^{(l)}+1}$ is the output vector of layer $l$, $M^{(l)}$ is the number of neurons in this layer, and $z_0^{(l)}$ corresponds to the output of a bias neuron fixed at one. $w_{ji}^{(l)}$ denotes the weight connecting neuron $i$ in layer $l-1$ to neuron $j$ in layer $(l)$. The output for neuron $j$ is then given by
\begin{equation}
z_j^{(l)} = g(a_j^{(l)}).
\end{equation}
Finally, the output for neuron $k$ in the output layer $(L+1)$ is similarly computed as:
\begin{equation}
y_k = f(a_k^{(L+1)})= f(\mathbf{w}_k^{(L+1)\mathrm{T}} \mathbf{z}^{(L)})
\end{equation}
where $f(\cdot)$ is the activation function for the output neurons, and $\mathbf{w}_k^{(L+1)}$ is the weight connecting the neurons in the final hidden layer $(L)$ to output neuron $k$.

Training an MLP seeks the optimal weights $\mathbf{w}$ that minimize a predetermined loss function $E(\mathbf{w})$, such as mean squared error or cross-entropy loss. Gradient descent is a widely used optimization method, in which weights are iteratively updated in the direction that locally reduces the value of $E(\mathbf{w})$. To efficiently compute gradients necessary for this optimization, the backpropagation algorithm employs the chain rule, propagating the error signals sequentially from the output layer backward through each hidden layer to the input layer.

Consider an output neuron $k$ in the output layer $(L+1)$. Using the chain rule, the partial derivative of the loss function $E(\mathbf{w})$ with respect to the weight $w_{kj}^{(L+1)}$ connecting neuron $j$ in the last hidden layer $(L)$ to neuron $k$ in the output layer is computed as follows:
\begin{equation}
\frac{\partial E(\mathbf{w})}{\partial w_{kj}^{(L+1)}} = \frac{\partial E(\mathbf{w})}{\partial a_k^{(L+1)}} \frac{\partial a_k^{(L+1)}}{\partial w_{kj}^{(L+1)}} = \delta_k^{(L+1)} z_j^{(L)}
\end{equation}
where
\begin{equation}
\delta_k^{(L+1)} \equiv \frac{\partial E(\mathbf{w})}{\partial a_k^{(L+1)}}=f'(a_k^{(L+1)})\frac{\partial E(\mathbf{w})}{\partial y_k}.
\end{equation}
This quantity, often referred to as the error for output neuron $k$, is the partial derivative of the loss $E(\mathbf{w})$ with respect to the neuron's activation $a_k^{(L+1)}$. By the chain rule, this derivative can be factored into two components: the partial derivative of the loss function with respect to the neuron's output, $\frac{\partial E(\mathbf{w})}{\partial y_k}$, and the derivative of the activation function evaluated at $a_k^{(L+1)}$, namely $f'(a_k^{(L+1)})$.

To calculate the error $\delta_j^{(l)}$ for hidden layer $(l)$, we make use of the chain rule again. The partial derivative of the loss function with respect to the weight $w_{ji}^{(l)}$ is:
\begin{equation}
 \frac{\partial E(\mathbf{w})}{\partial w_{ji}^{(l)}} = \frac{\partial E(\mathbf{w})}{\partial a_j^{(l)}}\frac{\partial a_j^{(l)}}{\partial w_{ji}^{(l)}} = \delta_j^{(l)} z_i^{(l-1)}.
\end{equation}
The error $\delta_j^{(l)}$ is defined as:
\begin{equation}
 \delta_j^{(l)} \equiv \frac{\partial E(\mathbf{w})}{\partial a_j^{(l)}} = g'(a_j^{(l)}) \sum_{k}\delta_k^{(l+1)} w_{kj}^{(l+1)},
\end{equation}
where $g'(\cdot)$ represents the derivative of the activation function $g(\cdot)$ evaluated at $a_j^{(l)}$. Intuitively, the error at hidden neuron $j$ is computed based on the weighted sum of error signals from the neurons in the subsequent layer $(l+1)$, thus sequentially propagating the error signal backward through the network.

Once the partial derivatives $\frac{\partial E(\mathbf{w})}{\partial w_{ji}^{(l)}}$ are computed for all weights, the update rule in standard gradient descent is:
\begin{equation}
w_{ji}^{(l)} \leftarrow w_{ji}^{(l)} - \eta \frac{\partial E(\mathbf{w})}{\partial w_{ji}^{(l)}},
\end{equation}
where $\eta$ is the learning rate. This iterative cycle—consisting of a forward pass (computation of neuron activations), a backward pass (error computation), and weight updates—continues until a predetermined convergence criterion is satisfied or a specified number of training epochs has been reached.

The gradient of the loss function with respect to the weights is computed using training data. In practice, to compute the gradient, most practitioners use a method called stochastic gradient descent (SGD) \citep{Lecun:2015}. In SGD, the calculation of the gradient is performed on a subset of the training data, known as a mini-batch. This approach enables efficient computation of the gradient, making it feasible to train large neural networks using extensive datasets.

Traditional MLPs have an advantage when handling network architectures requiring multiple outputs. MLPs can simultaneously optimize the weights associated with each output. This is attributed to the backpropagation mechanism in MLPs, where the weights connecting the output layer to the preceding hidden layer are specifically adjusted to minimize the error associated with each output. Concurrently, hidden-layer neurons are trained to extract and generalize pertinent features from the input data. This hierarchical optimization enables MLPs to efficiently model complex input-output relationships and successfully generalize the learned features.

\section*{Supplementary Results}

\section*{Overview}
This document provides supplementary results and details omitted from the main manuscript for brevity. Unless stated otherwise, experimental settings follow those in the main text.

\section{Additional parity-check results for $n_{\mathrm{bit}}\in\{2,3,4,5\}$}
\label{sec:supp-parity}

In this supplementary, we report full results for $n_{\mathrm{bit}}\in\{2,3,4\}$, and we also include $n_{\mathrm{bit}}=5$ in the plots for direct comparison with the main text.


To evaluate the efficacy of the Error Diffusion Learning Algorithm (EDLA), we initially assessed its performance on a parity check task, a classic benchmark in neural network learning. This task involves determining whether the number of ones in a binary input vector is even or odd. The parity check task is a two-class classification problem. The EDLA network consists of input, hidden, and output layers. The neurons in the hidden layers use the sigmoid or ReLU activation functions (AFs). The network has a single-output neuron that employs the sigmoid AF, producing a value between 0 and 1. The output is interpreted as the probability of the input vector containing an odd number of ones. Thus, the output neuron is expected to yield a value above 0.5 for odd counts and below 0.5 for even counts. The total number of possible input combinations is $2^{n_{\mathrm{bit}}}$, where $n_\mathrm{bit}$ denotes the number of bits in the input vector. In this experiment, we consider $n_{\mathrm{bit}}\in\{2,3,4,5\}$, and each dataset contains all $2^{n_{\mathrm{bit}}}$ input combinations. A mini-batch size of 4 is consistently employed across all configurations.

\begin{figure}[htbp]
 \centering
 \includegraphics[width=0.8\linewidth]{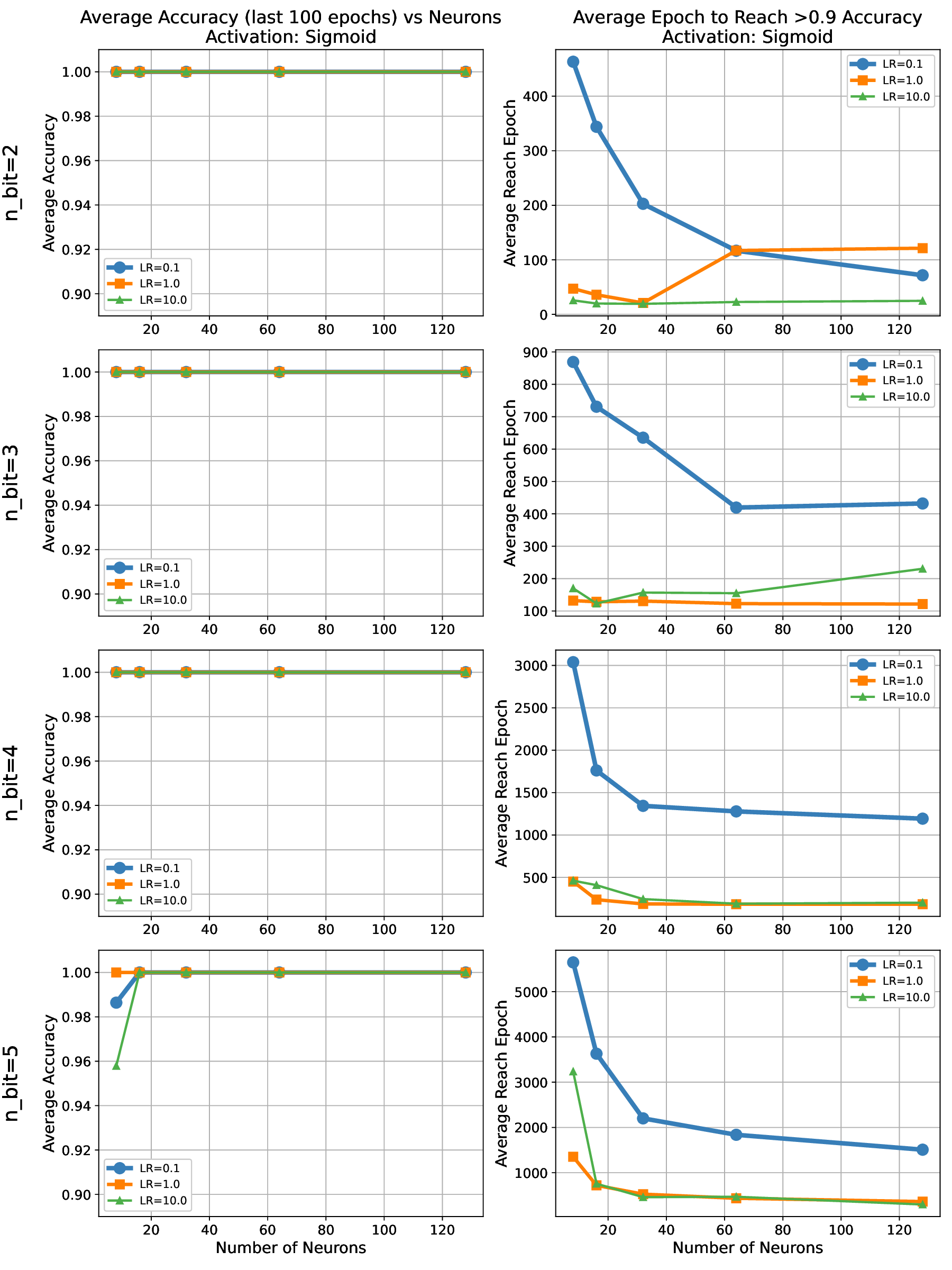}
 \caption{Performance of EDLA with the sigmoid AF across various learning rates on the parity check task. The left column depicts the accuracy of the EDLA network with the sigmoid AF as a function of neurons in a single hidden layer for different learning rates (LR = 0.1, 1.0, 10.0). Unless stated otherwise, parity-check experiments are run for 20,000 epochs; reported accuracies are averaged over the last 100 epochs. The right column shows the number of epochs required to reach an accuracy of 0.9 as a function of the number of neurons in a hidden layer for the same learning rates. Each row of plots corresponds to a different bit length (n\_bit, ranging from 2 to 5), for the parity check task. All results are averaged over 10 independent trials using specific random seeds (48835, 52642, 7841, 58416, 96828, 34439, 25155, 52094, 23535, 49704).}
 \label{fig:edla_sigmoid_parity_accu_epoch_lr}
\end{figure}

Figure \ref{fig:edla_sigmoid_parity_accu_epoch_lr} illustrates the performance of the EDLA network with a single hidden layer and the sigmoid AF across various learning rates (0.1, 1.0, 10.0) and neuron counts per layer (8, 16, 32, 64, 128). The results demonstrate that the EDLA network achieves high accuracy on the parity check task. The network attains near-perfect accuracy in almost all settings; the only noticeable degradations occur at the smallest width (8 neurons) for $n_{\mathrm{bit}}=5$, where performance depends on the learning rate. The epochs required to reach an accuracy of 0.9 generally decrease as the number of neurons increase, although some settings (e.g., $n_{\mathrm{bit}}=2$ at larger widths) show non-monotonic behavior. For learning rates of 1.0 and 10.0, convergence is generally faster than for a learning rate of 0.1 across most configurations. The relative ordering between LR=1.0 and LR=10.0 is generally close but can be width- and task-dependent. These results suggest that learning rates of 10 and 1.0 enable efficient training.

\begin{figure}[htbp]
 \centering
 \includegraphics[width=0.8\linewidth]{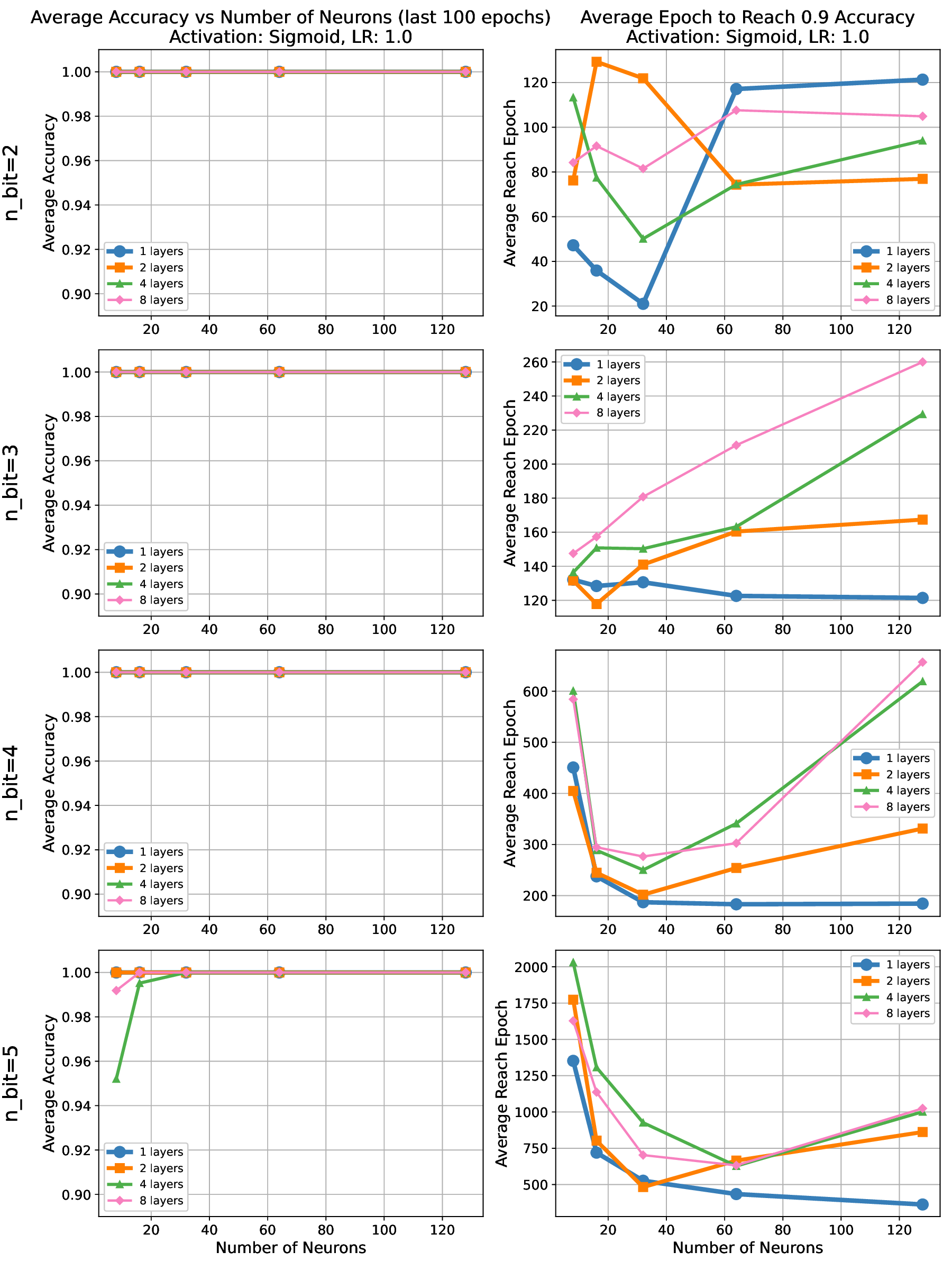}
 \caption{Accuracy and convergence characteristics of the EDLA network with the sigmoid AF across various network depths (1, 2, 4, and 8 hidden layers) at a fixed learning rate of 1.0. The left column shows the average classification accuracy as a function of the neurons per hidden layer. Unless stated otherwise, parity-check experiments are run for 20,000 epochs; reported accuracies are averaged over the last 100 epochs. The right column presents the number of epochs required to reach an accuracy of 0.9 as a function of the number of neurons per hidden layer. Each row of plots corresponds to different bit lengths (n\_bit, ranging from 2 to 5). All results are averaged over 10 independent trials, employing the same random seeds used in Fig.~\ref{fig:edla_sigmoid_parity_accu_epoch_lr} to ensure consistency.}
 \label{fig:edla_sigmoid_parity_accu_epoch}
\end{figure}

Figure \ref{fig:edla_sigmoid_parity_accu_epoch} presents an analysis of the performance of the EDLA network using the sigmoid AF across various hidden layers (1, 2, 4, and 8 hidden layers) and neuron counts per layer with a fixed learning rate of 1.0. The results reveal that the accuracy consistently approaches or reaches 1.0 across different bit numbers ($n_{\mathrm{bit}}$), regardless of the number of hidden layers. However, the number of epochs required to achieve 0.9 accuracy reveals more nuanced dependencies on the network depth and neuron count. Specifically, for $n_{\mathrm{bit}}=2$, shallow networks (one layer) exhibit a relatively low reach epoch for smaller neuron numbers, but the reach epoch shows a non-monotonic dependence on width (fastest at intermediate widths). The deeper networks show weaker and less consistent reductions than the 1-layer case. Conversely, for $n_{\mathrm{bit}}\in\{3, 4\}$ with deeper architectures (four and eight layers), the reach epoch tends to increase significantly at larger neuron counts. For $n_{\mathrm{bit}} = 5$, the reach epoch decreases notably with the neuron count up to around 32 or 64 neurons, while a mild slowdown can appear at 128 neurons. These results indicate that EDLA networks with the sigmoid AF can consistently achieve high accuracy on the parity check task, with optimal performance depending on the complexity of the task and the chosen network configuration.

\begin{figure}[htbp]
 \centering
 \includegraphics[width=0.8\linewidth]{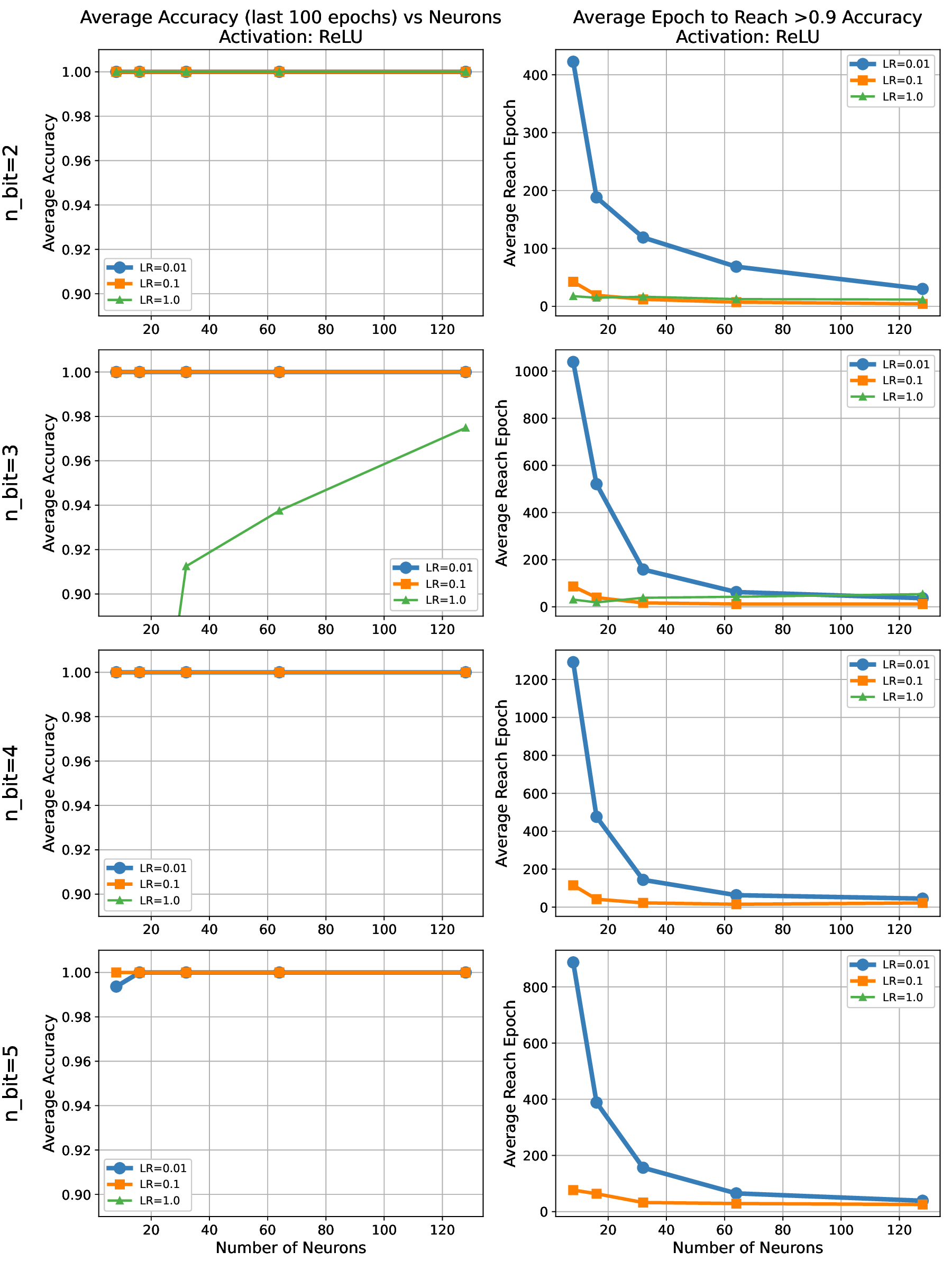}
 \caption{Performance of EDLA with the ReLU AF across various learning rates in the parity check task. The left column shows the accuracy of the EDLA network as a function of neurons in a single hidden layer for different learning rates (LR = 0.01, 0.1, 1.0). Unless stated otherwise, parity-check experiments are run for 20,000 epochs; reported accuracies are averaged over the last 100 epochs. The right column shows the number of epochs required to achieve an accuracy of 0.9 against the number of neurons in the hidden layer for the same learning rates. Each row of plots corresponds to a different bit length (n\_bit), ranging from 2 to 5, for the parity check task. Missing data points in the left column indicate that the accuracy is excessively low. Missing data points in the right column indicate that the accuracy does not reach 0.9. All results are averaged over 10 independent trials, employing the same random seeds used in Fig.~\ref{fig:edla_sigmoid_parity_accu_epoch_lr} to ensure consistency.}
 \label{fig:edla_relu_parity_accu_epoch_lr}
\end{figure}

Figure \ref{fig:edla_relu_parity_accu_epoch_lr} shows the performance of the EDLA network with a single hidden layer and the ReLU AF across various learning rates (0.01, 0.1, 1.0) and neuron counts of the hidden layer. For learning rates of 0.01 and 0.1, the network achieves high accuracy on the parity check task, with the number of epochs required to reach an accuracy of 0.9 generally decreasing as the number of neurons increases. For a learning rate of 0.01, the network exhibits a slower convergence speed than a learning rate of 0.1, particularly for smaller neuron counts. For a learning rate of 1.0, the performance is worse than that for the other learning rates in more complex tasks ($n_{\mathrm{bit}} \ge 3$), although it converges rapidly for the simplest task ($n_{\mathrm{bit}} = 2$). In particular, for LR = 1.0 and $n_{\mathrm{bit}} = 3$, small widths may transiently reach 0.9 but do not reliably maintain it, leading to low final accuracy at small widths, while for $n_{\mathrm{bit}} \in \{4, 5\}$, the network completely fails to achieve an accuracy of 0.9 even with a large number of neurons. This shows that a learning rate of 1.0, which is too high for the EDLA network with the ReLU in increasingly complex settings, leads to instability in training and poor performance.

\begin{figure}[htbp]
 \centering
 \includegraphics[width=0.8\linewidth]{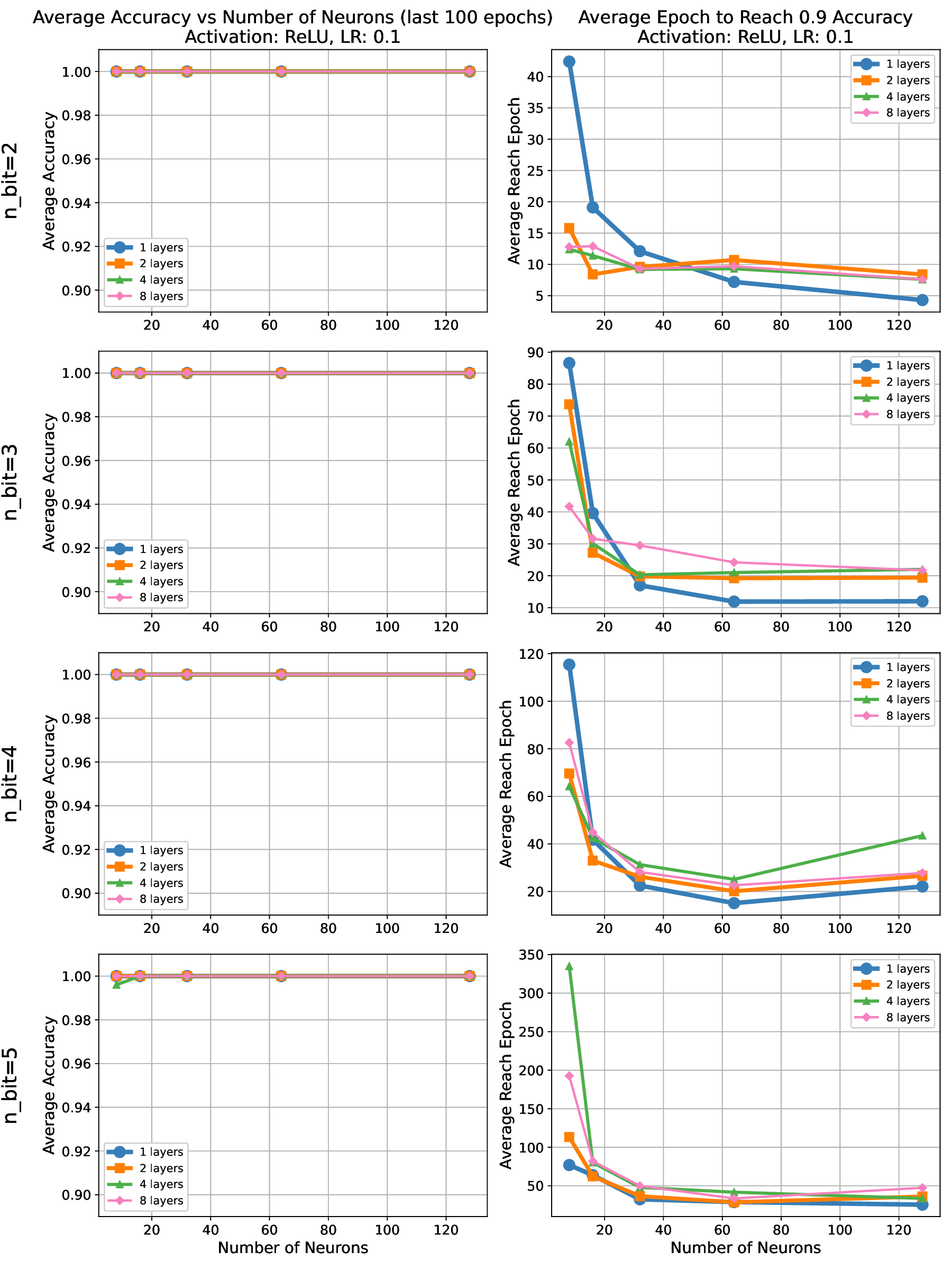}
 \caption{Accuracy and convergence characteristics of the EDLA network with ReLU AF across various network depths (1, 2, 4, and 8 hidden layers), using a fixed learning rate of 0.1. The left column shows average classification accuracy as a function of neurons per hidden layer. Unless stated otherwise, parity-check experiments are run for 20,000 epochs; reported accuracies are averaged over the last 100 epochs. The right column presents the number of epochs required to reach an accuracy of 0.9 as a function of the number of neurons per hidden layer. Each row of plots corresponds to different bit lengths (n\_bit), ranging from 2 to 5. All results are averaged over 10 independent trials, employing the same random seeds used in Fig.~\ref{fig:edla_sigmoid_parity_accu_epoch_lr} to ensure consistency.}
 \label{fig:edla_relu_parity_accu_epoch}
\end{figure}

Figure \ref{fig:edla_relu_parity_accu_epoch} shows the accuracy and convergence speed of EDLA with ReLU across neuron counts and hidden-layer depths at a fixed learning rate of 0.1. Across all $n_{\mathrm{bit}}$, accuracy saturates near 1.0 for most widths and depths. The reach epoch to 0.9 accuracy generally decreases as width increases, especially for shallow networks, but this trend is not monotonic and often exhibits diminishing returns. For deeper architectures (e.g., 4 and 8 layers), the reach epoch can display a U-shaped dependence on width, with a slowdown at the largest width (notably for $n_{\mathrm{bit}}=4$).

\end{document}